\documentclass[10pt,twocolumn,letterpaper]{article}
\usepackage{iccv}
\usepackage{times}
\usepackage{epsfig}
\usepackage{graphicx}
\usepackage{amsmath}
\usepackage{amssymb}
\usepackage{tabu}
\usepackage{subcaption}
\usepackage{multirow}
\usepackage{booktabs}
\usepackage{array}
\usepackage{titling}
\DeclareMathOperator{\sign}{sign}

\usepackage[pagebackref=true,breaklinks=true,letterpaper=true,colorlinks,bookmarks=false]{hyperref}

\newcommand{\norm}[1]{\left\lVert#1\right\rVert}

\newcommand{\thickhline}{%
    \noalign {\ifnum 0=`}\fi \hrule height 1pt
    \futurelet \reserved@a \@xhline
}
\newcolumntype{"}{@{\hskip\tabcolsep\vrule width 1pt\hskip\tabcolsep}}

 \iccvfinalcopy 

\def\iccvPaperID{2380} 

\ificcvfinal\fi

\begin{document}

\title{Robustness and Generalization via Generative Adversarial Training}

\author{Omid Poursaeed$^{1,2}$ \qquad Tianxing Jiang$^{1}$ \qquad Harry Yang$^{3}$\\ \\ {Serge Belongie}$^{1,2}$ \qquad {Ser-Nam Lim}$^{3}$ \\
    \newline  \\
	{$^1${Cornell University}\qquad}
    $^2${Cornell Tech}\qquad
    $^3${Facebook AI}\\	}

\date{}
\maketitle
\ificcvfinal\thispagestyle{empty}\fi



%


\begin{abstract}

While deep neural networks have achieved remarkable success in various computer vision tasks, they often fail to generalize to new domains and subtle variations of input images. 
Several defenses have been proposed to improve the robustness against these variations. 
However, current defenses can only withstand the specific attack used in training, and the models often remain vulnerable to other input variations. Moreover, these methods often degrade performance of the model on clean images and do not generalize to out-of-domain samples.  
In this paper we present \textit{Generative Adversarial Training}, an approach to simultaneously improve the model's generalization to the test set and out-of-domain samples as well as its robustness to unseen adversarial attacks. Instead of altering a low-level pre-defined aspect of images, we generate a spectrum of low-level, mid-level and high-level changes using generative models with a disentangled latent space. Adversarial training with these examples enable the model to withstand a wide range of attacks by observing a variety of input alterations during training. We show that our approach not only improves performance of the model on clean images and out-of-domain samples but also makes it robust against unforeseen attacks and outperforms prior work. 
We validate effectiveness of our method by demonstrating results on various tasks such as classification, segmentation and object detection. 
\end{abstract}

\section{Introduction}
Deep neural networks have shown promising generalization to in-domain samples. However, they are vulnerable to slight alterations of input images and have limited generalization to new domains.  
Several defenses have been proposed for improving the models' robustness against input variations. 
However, these models only provide robustness against a narrow range of threat models used in training, and they have poor generalization to unseen attacks. We hypothesize that this is due to the fact that they only consider a small subset of realistic examples on or near the manifold of natural images. For instance, additive perturbations leave high-level semantic aspects of images intact. Therefore, models trained against these examples do not provide robustness to high-level input variations. 
In order for a model to become robust to all realistic variations of inputs, it needs to see a diverse set of samples during training. 
However, most of the existing works only alter a low-level aspect of images such as color \cite{hosseini2018semantic,laidlaw2019functional,bhattad2019unrestricted}, spatial \cite{engstrom2017rotation,xiao2018spatially,alaifari2018adef}, pose \cite{alcorn2018strike,zeng2019adversarial}, and others. Even if we consider the union of several types of attacks, the model can still be vulnerable to other input variations that are not contained in any of the constituent threat models. To provide robustness against unforeseen attacks, we propose adversarial training against a range of low-level to high-level variations of inputs. We leverage generative models with disentangled latent representations to systematically build diverse and realistic examples without leaving the manifold of natural images. We show that our approach improves generalization and robustness of the model to unseen variations of input images \textit{without} training against any of them.

We build upon state-of-the-art generative models which disentangle factors of variation in images. We create fine and coarse-grained adversarial changes by manipulating various latent variables at different resolutions. Loss of the 
target network is used to guide the generation process. The pre-trained generative model constrains the search space for our adversarial examples to realistic images, thereby revealing the target model's vulnerability in the natural image space. We verify that we do not deviate from the space of realistic images with a user study as well as a t-SNE plot comparing distributions of real and adversarial images. As a result, we observe that including these examples in training the model enhances its accuracy on clean images as well as out-of-domain samples. Moreover, since the model has seen a variety of low-level and high-level alterations of images, it becomes robust to a wide range of adversarial examples including recoloring, spatial transformations, perceptual and additive perturbations.

Our contributions can be summarized as follows:
\begin{itemize}
    \item {We present \textit{Generative Adversarial Training (GAT)} to simultaneously improve the model's generalization to clean and out-of-domain samples and its robustness to \textit{unforeseen} attacks. Our approach is based on adversarial training with fine-grained unrestricted adversarial examples in which the attacker controls which aspects of the image to manipulate, resulting in a diverse set of realistic, on-the-manifold examples.  } 
    \item {We evaluate GAT against a diverse set of adversarial attacks: recoloring, spatial transformations, perceptual and additive perturbations}, and demonstrate that it achieves state-of-the-art robustness against these attacks \textit{without} training against any of them. 
    \item {We extend our approach to semantic segmentation and object detection tasks, and propose the first method for generating unrestricted adversarial examples for segmentation and detection. 
    Training with our examples improves both robustness and generalization of the model. } 
\end{itemize}

\section{Related Work}
\subsection{Adversarial Examples} 
Most of the existing works on adversarial attacks focus on norm-constrained adversarial examples: for a given classifier $F:\mathbb{R}^n \rightarrow \{1, \ldots , K\}$ and an image $x\in \mathbb{R}^n$, the adversarial image $x'\in \mathbb{R}^n$ is created such that $\norm{x-x'}_p < \epsilon$ and $F(x)\neq F(x')$. Common values for $p$ are $0, 2, \infty$, and $\epsilon$ is chosen small enough so that the perturbation is imperceptible. Various algorithms have been proposed for creating $x'$ from $x$.
Optimization-based methods solve a surrogate optimization problem based on the classifier's loss and the perturbation norm \cite{szegedy2013intriguing,fletcher2013practical,carlini2017towards}. 
%
Gradient-based methods use gradient of the classifier's loss with respect to the input image \cite{goodfellow2014explaining,madry2017towards,dong2018boosting,kurakin2016adversarial}. 

Another line of work creates unrestricted adversarial examples that are not bounded by a norm threshold.  
One way to achieve this is by applying subtle geometric transformations such as spatial transformations \cite{xiao2018spatially,alaifari2018adef}, translations and rotations \cite{engstrom2017rotation} or pose changes \cite{alcorn2018strike} to the inputs. Other works consider recoloring \cite{hosseini2018semantic,laidlaw2019functional,bhattad2019unrestricted}, intermediate features  \cite{dunn2020evaluating,laidlaw2020perceptual,xu2020towards} and inserting new objects or patches in the image \cite{brown2017adversarial}. 
A challenge for creating unrestricted adversarial examples and defending against them is introduced in \cite{brown2018unrestricted} using the simple task of classifying between birds and bicycles.  
%
%
Recent works consider using generative models to create adversarial examples. Song et al. \cite{song2018constructing} 
search in the latent ($z$) space of AC-GAN \cite{odena2017conditional} to find generated images that can fool a target classifier but yield correct predictions on AC-GAN's auxiliary classifier. They constrain the search region of $z$ so that it is close to a randomly sampled noise vector, and show results on MNIST, SVHN and CelebA datasets.   
This approach does not alter any aspects of images and merely generates a set of generated images misclassified by the model. As we show in the supplementary material, training with these adversarial examples hurts the classifier's performance on clean images. 
One reason for this accuracy drop is that requiring two classifiers to have inconsistent predictions degrades sample quality of the model. 
To further illustrate difference of our approach with \cite{song2018constructing}, we plot t-SNE embeddings of real CelebA-HQ images and adversarial examples from our method and \cite{song2018constructing} in the supplementary material, and show that our adversarial images stay closer to the manifold of real images.    
The recent work by \cite{gowal2020achieving} shows that adversarial training with examples generated by StyleGAN can improve performance of the model on clean images. 
Their approach requires precomputing a mapping from the image space to the latent space for the whole dataset, which is computationally prohibitive for large datasets. It
is also constrained to fine changes using only a subset of the latent variables. They argue that coarse changes might be label-dependent. This statement is not true since coarse stylistic changes do not alter the label (e.g. gender) and merely modify high-level aspects of images. Moreover, \cite{gowal2020achieving} only considers the classification task on low-resolution datasets such as ColorMNIST (28x28) and CelebA (64x64). While their approach uses the StyleGAN model, they do not show any results on high-resolution datasets that StyleGAN is originally trained on (e.g. LSUN and CelebA-HQ (1024x1024)), which makes it hard to ascertain that their adversarial training will be effective on high-resolution datasets. Even on low-resolution datasets such as Color-MNIST, their adversarial training can perform worse than random sampling on unbiased datasets as shown in Table 2 of \cite{gowal2020achieving}. On the other hand, our approach directly samples and manipulates latent variables without requiring the mapping step. 
We demonstrate that by limiting the number of iterations we can use both coarse and fine changes and both contribute to improvements in performance. 
While \cite{gowal2020achieving} only enhances the accuracy on clean images, our approach also improves generalization to out-of-domains samples and unseen adversarial examples.  
In addition, we propose the first method for unrestricted adversarial attacks on semantic segmentation and object detection, and demonstrate that adversarial training improves segmentation and detection results on clean images. 

\subsection{Adversarial Robustness}
Several methods have been proposed for defending against adversarial attacks. 
Many defenses attempt to combat adversaries using a form of input pre-processing or by manipulating intermediate features or gradients \cite{guo2017countering,xie2017mitigating,samangouei2018defense}. Few approaches have been able to scale up to high-resolution datasets such as ImageNet \cite{liao2018defense,xie2018feature,kannan2018adversarial}.
Most of the proposed heuristic defenses were later broken by stronger adversaries \cite{carlini2017towards,uesato2018adversarial,athalye2018obfuscated}.
One of the most successful defenses is \textit{adversarial training} \cite{goodfellow2014explaining,kurakin2016adversarial,madry2017towards,xie2020adversarial,stutz2019disentangling,gowal2020achieving} which augments training data with adversarial examples generated as the training progresses. This approach is able to withstand strong attacks. 
Adversarial training with perturbation-based examples degrades performance of the model on clean images. \cite{xie2020adversarial} proposes to use separate batch norm layers for clean and adversarial images to avoid the accuracy drop.  
Our approach improves the model's accuracy on clean images without modifying the architecture. 
Moreover, unlike prior work which make the classifier robust only against the specific attack used in training, our method provides generalizable robustness across a range of attacks. 


\vspace{-0.1cm}
\section{Approach}\label{sec:approach}
We present \textit{Generative Adversarial Training}, a method for improving generalization and robustness of models to unseen attacks. Most of the existing works on adversarial attacks
modify a low-level aspect of images, thereby a model adversarially trained with these samples remains susceptible to various other input alterations. 
We demonstrate that by encouraging diversity and realism in the generated adversarial examples, we can improve both performance of the model on clean and out-of-domain images and its robustness to a wide range of adversarial attacks. Our approach creates a spectrum of low-level, mid-level and high-level changes for which the target network fails to generalize. The adversarially trained model observes a variety of examples on or near the manifold of natural images. This allows the model to generalize better to test samples and various alterations of images.   
We leverage disentangled latent representations for generating the adversarial examples. We build upon state-of-the-art generative models and use StyleGAN \cite{karras2018style} for the classification task and SPADE \cite{park2019semantic} for semantic segmentation and object detection.  

\begin{figure*}
\begin{center}
   \includegraphics[width=0.7\linewidth]{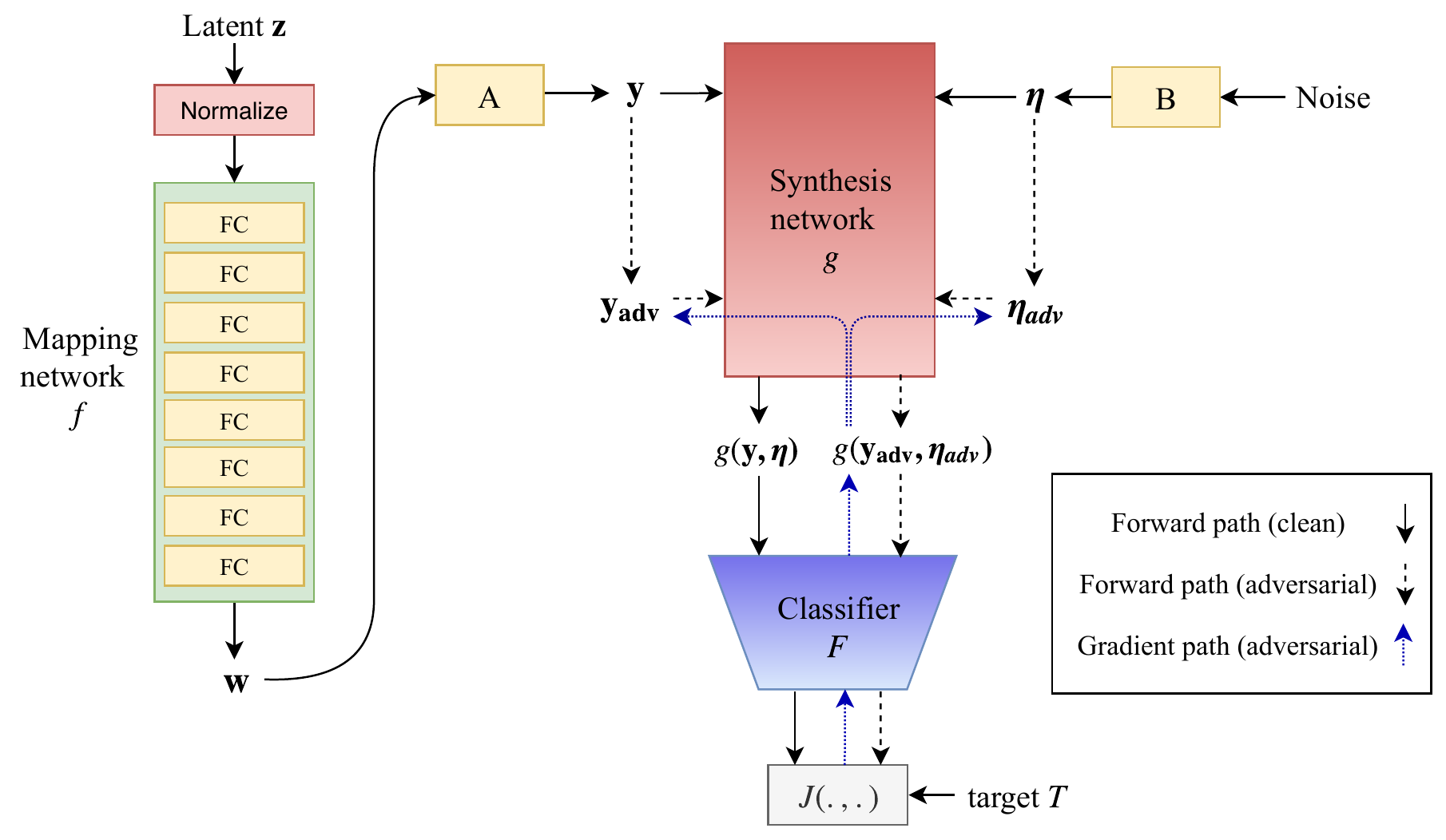}
\end{center}
\vspace{-0.4cm}
   \caption{Classification architecture. Style $({\bf{y}})$ and noise $(\boldsymbol{\eta})$ variables are used to generate images $g({\bf y}, \boldsymbol{\eta})$ which are fed to the classifier $F$.  
   Adversarial style and noise tensors are initialized with ${\bf y}$ and $\boldsymbol{\eta}$ and iteratively updated using gradients of the loss function $J$. The classifier $F$ is adversarially trained with clean and adversarial samples.} 
   \label{fig:architecture}
\vspace{-0.2cm}
\end{figure*}

\subsection{Classification} 
Style-GAN \cite{karras2018style} is a state-of-the-art generative model which disentangles high-level attributes and stochastic variations in an unsupervised manner. 
Stylistic variations are represented by \textit{style} variables and stochastic details are captured by \textit{noise} variables. 
 Changing the noise only affects low-level details, leaving the overall composition and high-level aspects  intact. This allows us to manipulate the noise variables such that variations are barely noticeable by the human eye. 
 The style variables affect higher level aspects of image generation. 
 For instance, when the model is trained on bedrooms, style variables from the top layers control viewpoint of the camera, middle layers select the particular furniture, and bottom layers deal with colors and details of materials \cite{karras2018style}.
 This allows us to manipulate images in a controlled manner, providing an avenue for fine-grained unrestricted attacks.  
 
Formally, we can represent Style-GAN with a mapping function $f$ and a synthesis network $g$. As illustrated in Figure \ref{fig:architecture}, the mapping function is an 8-layer MLP which takes a latent code ${\bf z}$, and produces an intermediate latent vector ${\bf w} = f({\bf z})$. This vector is then specialized by learned affine transformations $A$ to style variables ${\bf y}$, which control adaptive instance normalization operations after each convolutional layer of the synthesis network $g$. Noise inputs are single-channel images consisting of un-correlated Gaussian noise that are fed to each layer of the synthesis network. Learned per-feature scaling factors $B$ are used to generate noise variables $\boldsymbol{\eta}$ which are added to the output of convolutional layers.  
The synthesis network takes style ${\bf y}$ and noise $\boldsymbol{\eta}$ as input, and generates an image ${\bf x} = g({\bf y}, \boldsymbol{\eta})$. 
We pass the generated image to a pre-trained classifier $F$. 
We seek to slightly modify ${\bf x}$ so that $F$ can no longer classify it correctly. We achieve this through perturbing the style and noise tensors. 
We initialize adversarial style and noise variables as ${\bf y_{adv}^{(0)}=y}$ and $\boldsymbol{\eta}_{\bf adv}^{\bf (0)}=\boldsymbol{\eta}$, and iteratively update them in order to fool the classifier. Loss of the classifier determines the update rule, which in turn depends on the type of attack.  
As common in the literature, we consider two types of attacks: non-targeted and targeted.     

In order to generate non-targeted adversarial examples, we need to change the model's original prediction. Starting from initial values ${\bf y_{adv}^{(0)}=y}$ and $\boldsymbol{\eta}_{\bf adv}^{\bf (0)}=\boldsymbol{\eta}$, we can iteratively perform gradient ascent in the style and noise spaces of the generator to find values that maximize the classifier's loss. Alternatively, as proposed by \cite{kurakin2016adversarial}, we can use the least-likely predicted class $ll_{\bf{x}}=\arg\min(F({\bf x}))$ as our target. We found this approach more effective in practice.
At time step $t$, the update rule for style and noise variables is: 

\begin{equation}\label{eq:style-ll}
    {\bf {y_{adv}^{(t+1)}}} = {\bf y_{adv}^{(t)}} - \epsilon \cdot \sign (\nabla_{\bf y_{adv}^{(t)}}J(F(g({\bf y_{adv}^{(t)}}, \boldsymbol{\eta}_{\bf adv}^{\bf (t)})), ll_{\bf x}))
\end{equation}
\begin{equation}\label{eq:noise-ll}
    {\boldsymbol{\eta}_{\bf adv}^{\bf(t+1)}} = \boldsymbol{\eta}_{\bf adv}^{\bf(t)} - \delta \cdot \sign (\nabla_{\boldsymbol{\eta}_{\bf adv}^{\bf(t)}}J(F(g({\bf y_{adv}^{(t)}}, \boldsymbol{\eta}_{\bf adv}^{\bf(t)})), ll_{\bf x}))
\end{equation}
in which $J(\cdot, \cdot)$ is the classifier's loss function, $F(\cdot)$ gives the probability distribution over classes, 
${\bf x} = g({\bf y}, \boldsymbol{\eta})$, and $\epsilon, \delta \in \mathbb{R}$ are step sizes. 
 We use $(\epsilon, \delta)=(0.004, 0.2)$ and $(0.004, 0.1)$ for LSUN and CelebA-HQ respectively. We perform multiple steps of gradient descent (usually 2 to 10) until the classifier is fooled.
 We report the average number of iterations required to fool the classifier in the supplementary material. 




Generating targeted adversarial examples is more challenging as we need to change the prediction to a specific class $T$. In this case, 
we perform gradient descent to minimize the classifier's loss with respect to the target $T$:
\begin{equation}\label{eq:targeted-style}
    {\bf y_{adv}^{(t+1)}} = {\bf y_{adv}^{(t)}} - \epsilon \cdot \sign (\nabla_{\bf y_{adv}^{(t)}}J(F(g({\bf y_{adv}^{(t)}}, \boldsymbol{\eta}_{\bf adv}^{\bf (t)})), T)) 
\end{equation}
\begin{equation}\label{eq:targeted-noise}
    {\boldsymbol{\eta}_{\bf adv}^{\bf (t+1)}} = \boldsymbol{\eta}_{\bf adv}^{\bf(t)} - \delta \cdot \sign (\nabla_{\boldsymbol{\eta}_{\bf adv}^{\bf (t)}}J(F(g({\bf y_{adv}^{(t)}}, \boldsymbol{\eta}_{\bf adv}^{\bf (t)})), T)) 
\end{equation}
We use $(\epsilon, \delta) = (0.005, 0.2)$ and $(0.004, 0.1)$ in the experiments on LSUN and CelebA-HQ respectively. 
Note that we only control deviation from the initial latent variables, and do not impose any norm constraint on generated images. 

The classifier $F$ is adversarially trained with equal number of clean and adversarial images. 
To maximize diversity of generated samples, we manipulate groups of consecutive style and noise layers separately (e.g. layers 1-2, 3-4, etc.) for experiments on adversarial training.   
As the training progresses, the classifier observes a variety of examples and becomes more robust to input variations. Hence, the generator needs to explore new areas of the natural image manifold that correspond to generalization errors of the classifier. Since our model allows both coarse and fine changes, it results in superior generalization performance compared to existing works on adversarial training that only manipulate a low-level aspect of images.

\begin{figure*}
\begin{center}
   \includegraphics[width=0.6\linewidth]{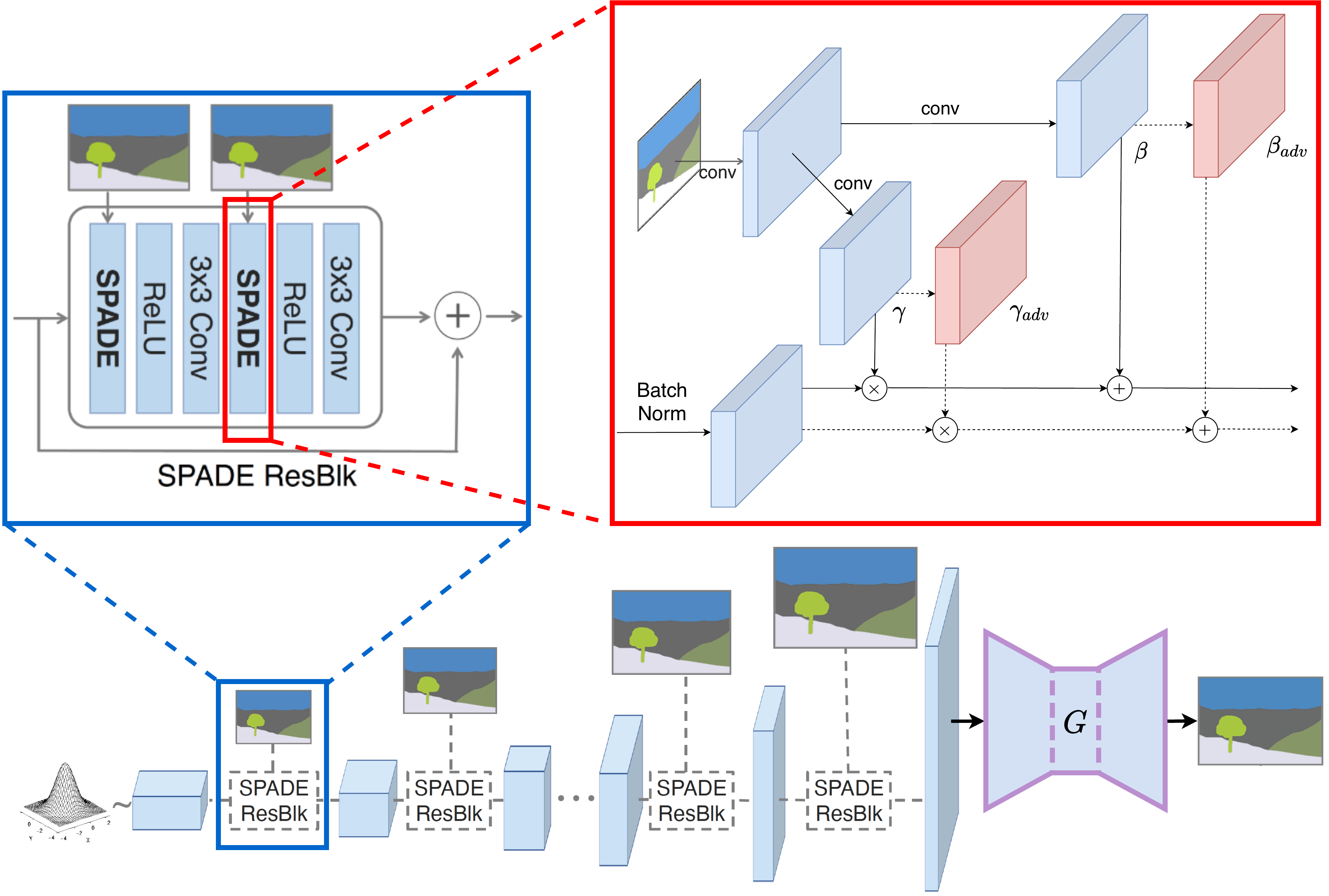}
\end{center}
\vspace{-0.5cm}
   \caption{\textcolor{black}{Semantic segmentation and object detection architecture. Adversarial parameters $\gamma_{adv}$ and $\beta_{adv}$ are initialized with $\gamma$ and $\beta$, and iteratively updated to fool the segmentation or detection model $G$. The model $G$ is adversarially trained with clean and adversarial examples. }  }\label{fig:segmentation}
\vspace{-0.3cm}
\end{figure*}

\subsubsection{Input-conditioned Generation }\label{sec:input-cond}
 Generation can also be conditioned on real input images by embedding them into the latent space of Style-GAN. 
  We first synthesize images similar to the given input image $I$ by optimizing values of ${\bf y}$ and $\boldsymbol{\eta}$ such that $g({\bf y}, {\boldsymbol{\eta}})$ is close to $I$. More specifically, we minimize the perceptual distance \cite{johnson2016perceptual} between $g({\bf y}, {\boldsymbol{\eta}})$ and $I$. 
  We can then proceed similar to equations {1--4} to perturb these tensors and generate the adversarial image. Realism of synthesized images depends on inference properties of the generative model. In practice, generated images resemble input images with high fidelity especially for CelebA-HQ images.  


\subsection{Semantic Segmentation and Object Detection}
\textcolor{black}{We also consider the task of semantic segmentation and leverage the generative model proposed by \cite{park2019semantic}. The model is conditioned on input semantic layouts and uses SPatially-Adaptive (DE)normalization (SPADE) modules to better preserve semantic information against common normalization layers. The layout is first projected onto an embedding space and then convolved to produce the modulation parameters $\gamma$ and $\beta$. We adversarially modify these parameters with the goal of fooling a segmentation model. 
We consider non-targeted attacks using per-pixel predictions and compute gradient of the loss function with respect to the modulation parameters with an update rule similar to equations \ref{eq:style-ll} and \ref{eq:noise-ll}. 
Figure~\ref{fig:segmentation} illustrates the architecture. 
We consider a similar architecture for the object detection task except that we pass the generated image to the detection model and try to increase its loss. Results for this task are shown in the supplementary material.    
} 

\section{Results and Discussion}
We provide qualitative and quantitative results using experiments on LSUN \cite{yu2015lsun} and CelebA-HQ \cite{karras2017progressive}. 
LSUN contains 10 scene categories 
and 20 object categories. We use all the scene classes as well as two object classes: \textit{cars} and \textit{cats}. 
We consider this dataset since it is used in Style-GAN, and is well suited for a classification task. 
For the scene categories, a 10-way classifier is trained based on Inception-v3 \cite{szegedy2016rethinking} which achieves an accuracy of ${88.9\%}$ on LSUN's test set. 
The two object classes also appear in ImageNet \cite{deng2009imagenet}, a richer dataset containing $1000$ categories. Therefore, for experiments on cars and cats we use an Inception-v3 model trained on ImageNet. This allows us to explore a broader set of categories in our attacks, and is particularly helpful for targeted adversarial examples.  
CelebA-HQ 
consists of 30,000 face images at $1024 \times 1024$ resolution. We consider the gender classification task, and use the classifier provided by \cite{karras2018style}. This is a binary task for which targeted and non-targeted attacks are similar. 

\begin{figure*}
\begin{subfigure}{1.0\textwidth}
\hspace{3.1cm}\includegraphics[width=0.75\linewidth]{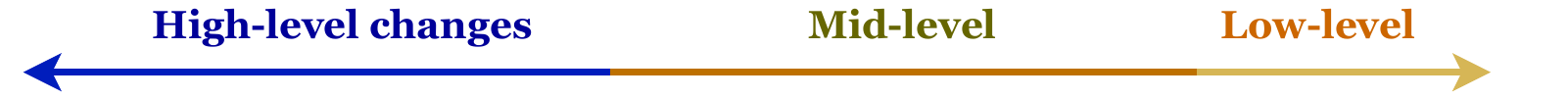}
\hspace{-0.2cm}
\vspace{-0.3cm}
\begin{center}
\includegraphics[width=0.8\linewidth]{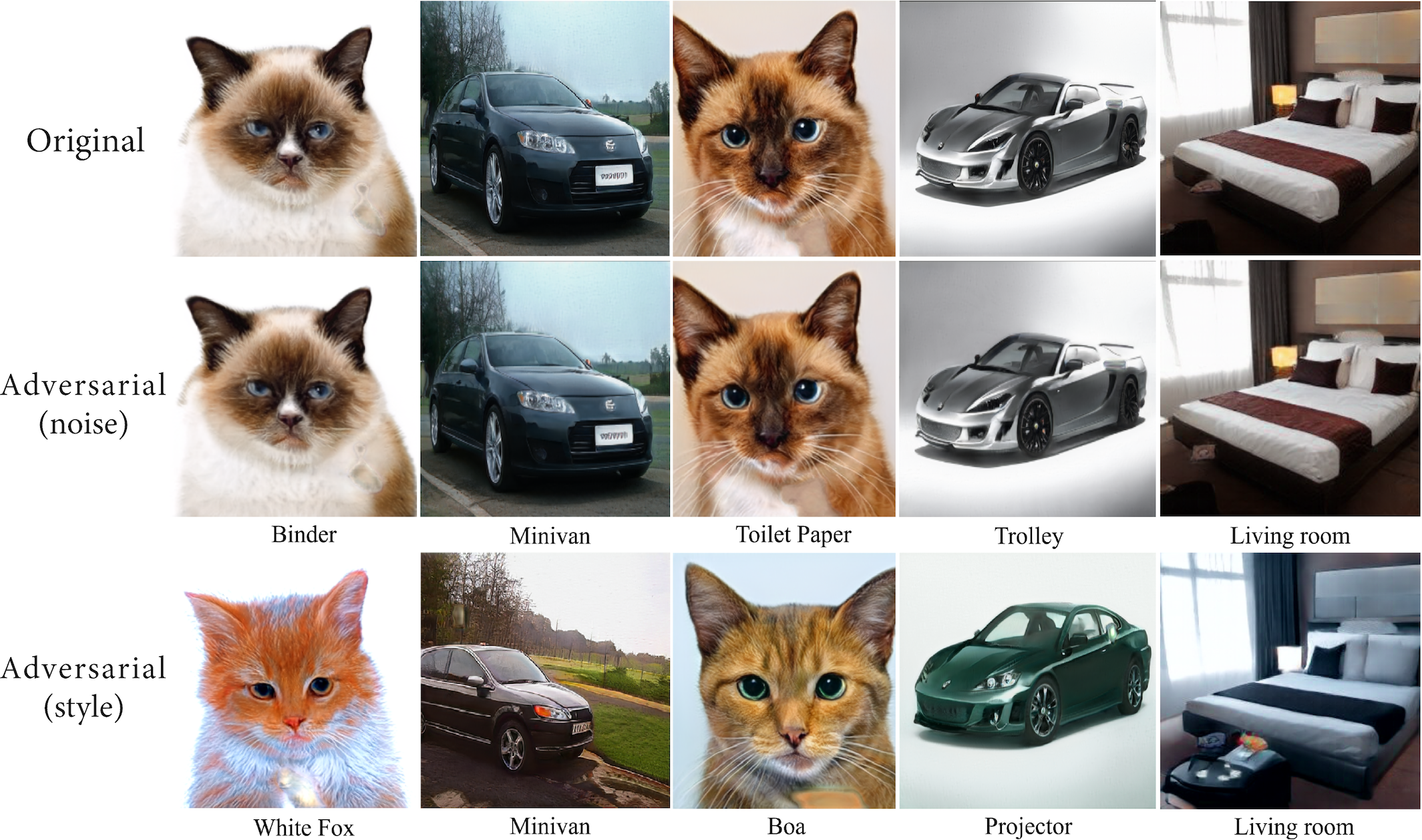}    
\end{center}
\vspace{-0.3cm}
\caption{Non-targeted}
\vspace{+0.1cm}
\end{subfigure}
\begin{subfigure}{1.0\textwidth}
\centering
\begin{center}
\includegraphics[width=0.8\linewidth]{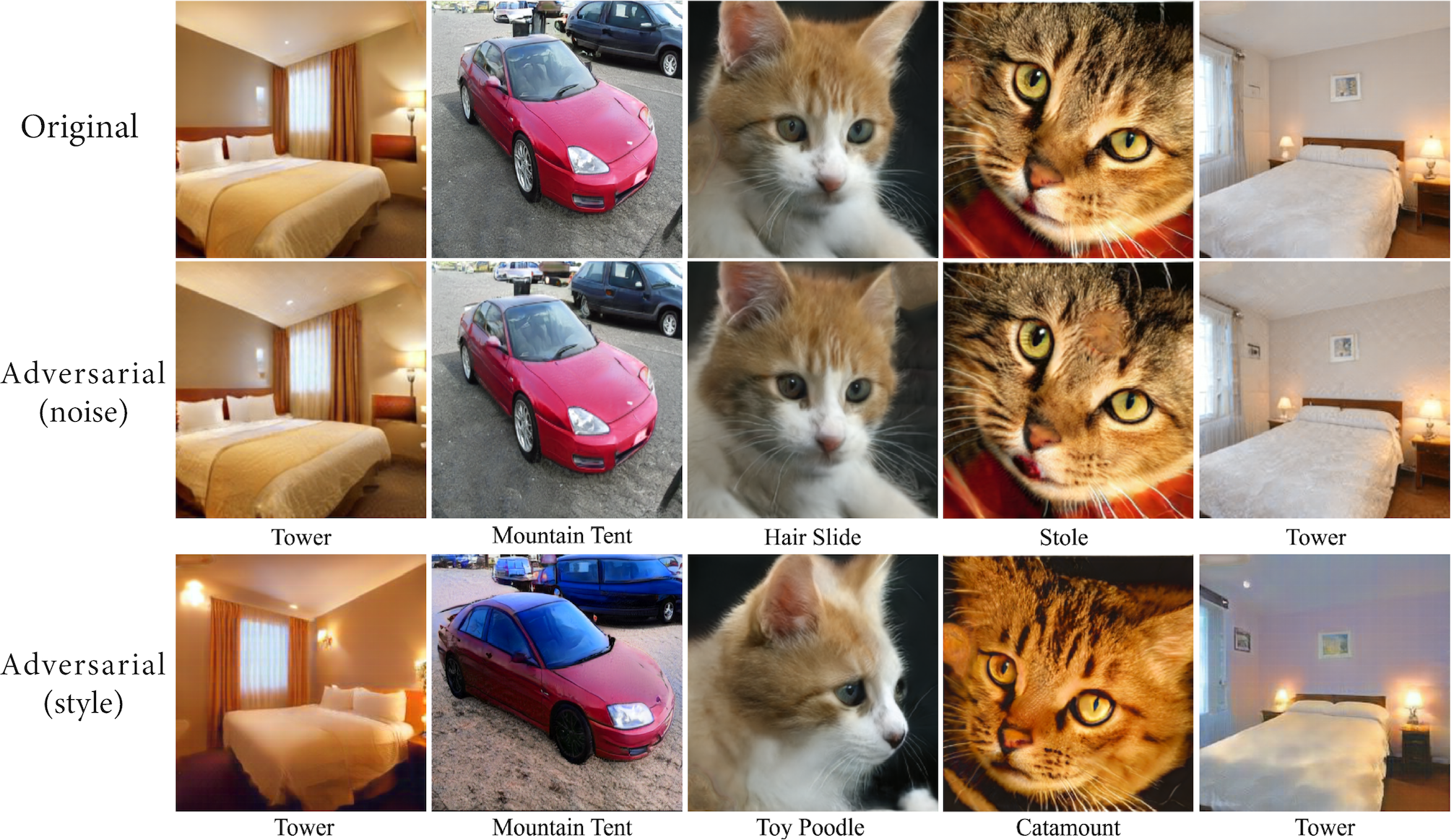}
\end{center}
\vspace{-0.4cm}
\caption{Targeted}
\end{subfigure}
\vspace{-0.2cm}
  \caption{Unrestricted adversarial examples on LSUN for a) non-targeted and b) targeted attacks. Predicted classes are shown under each image. 
  }
  \vspace{-0.4cm}
\label{fig:imgs}
\end{figure*}

In order to synthesize a variety of adversarial examples, we use different random seeds in Style-GAN to obtain various values for ${\bf z}, {\bf w}, {\bf y}$ and $\boldsymbol{\eta}$. Style-based adversarial examples are generated by initializing ${\bf y_{adv}}$ with the value of ${\bf y}$, and iteratively updating it as in equation \ref{eq:style-ll} (or \ref{eq:targeted-style}) until the resulting image $g({\bf y}_{\bf adv}, \boldsymbol{\eta})$ fools the classifier $F$. Noise-based adversarial examples are created similarly using $\boldsymbol{\eta}_{\bf adv}$ and the update rule in equation \ref{eq:noise-ll} (or \ref{eq:targeted-noise}). While using different step sizes makes a fair comparison difficult, we generally found it easier to fool the model by manipulating the noise variables. We can also combine the effect of style and noise by simultaneously updating ${\bf y}_{\bf adv}$ and  $\boldsymbol{\eta}_{\bf adv}$ in each iteration, and feeding $g({\bf y}_{\bf adv}, \boldsymbol{\eta}_{\bf adv})$ to the classifier. In this case, the effect of style usually dominates since it creates coarser changes. 
 

\begin{figure*}
\centering
\includegraphics[width=0.85\linewidth]{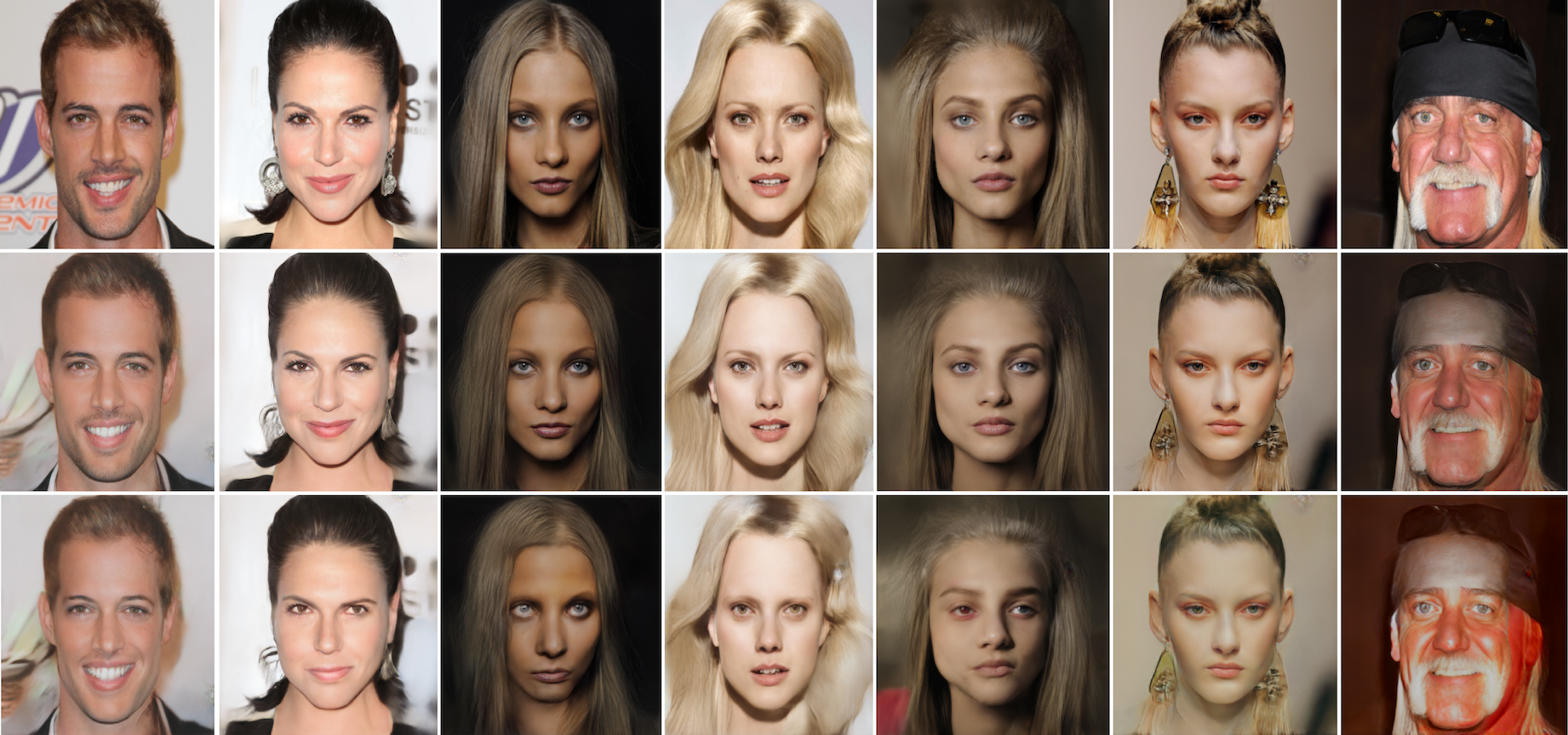}
\vspace{-0.1cm}
  \caption{Input-conditioned adversarial examples on CelebA-HQ gender classification. From top to bottom: input, generated and style-based images. Males are classified as females and vice versa. 
  }
  \vspace{-0.1cm}
\label{fig:input-cond}
\end{figure*}

\begin{figure*}
\centering
\includegraphics[width=0.85\linewidth]{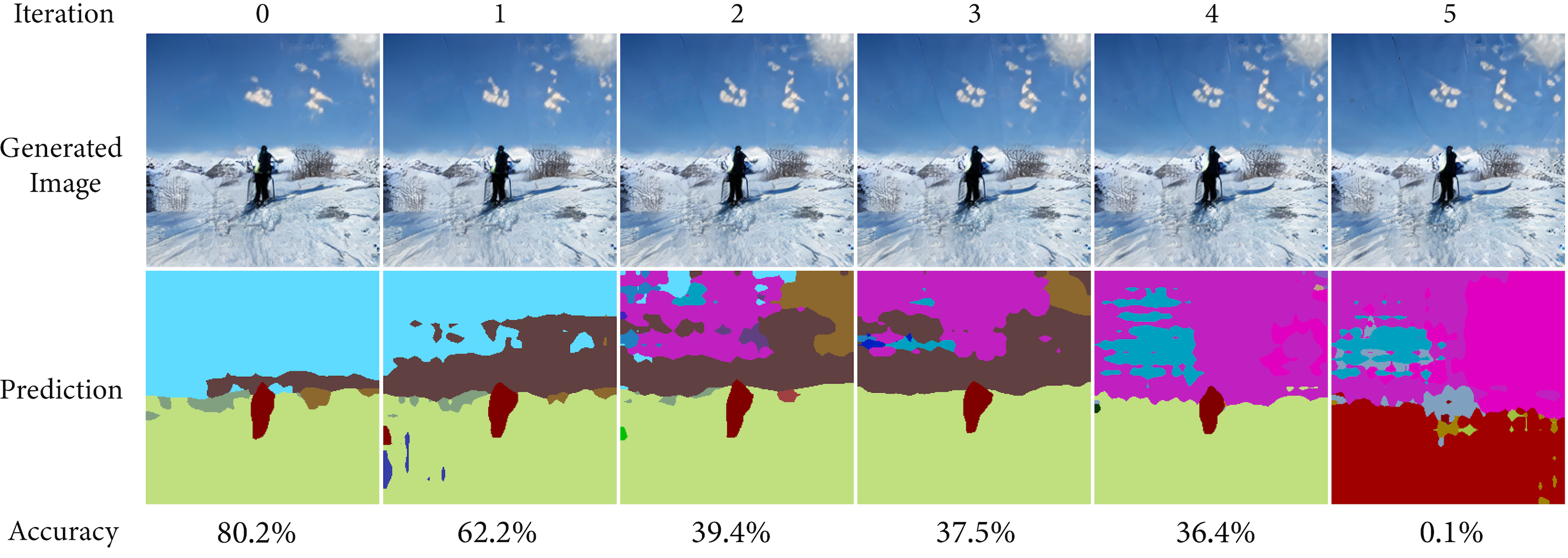}
\vspace{-0.2cm}
  \caption{\textcolor{black}{Unrestricted adversarial examples for semantic segmentation. Generated images, corresponding predictions and their accuracy (ratio of correctly predicted pixels) are shown for different number of iterations. }  
  }
\label{fig:seg-results}
\end{figure*}

Figure \ref{fig:imgs} illustrates generated adversarial examples 
on LSUN. Original images $g({\bf y}, \boldsymbol{\eta})$, noise-based images $g({\bf y}, \boldsymbol{\eta}_{\bf adv})$ and style-based images $g({\bf y}_{\bf adv}, \boldsymbol{\eta})$ are shown. 
Adversarial images look almost indistinguishable from natural images. 
Manipulating the noise variable results in subtle, imperceptible changes. 
Varying the style leads to coarser changes such as different colorization, pose changes, and even removing or inserting objects in the scene. 
{We can also control granularity of changes by selecting specific layers of the model. Manipulating top layers, corresponding to coarse spatial resolutions, results in high-level changes. Lower layers, on the other hand, modify finer details. In the first two columns of Figure \ref{fig:imgs}, we only modify top 6 layers (out of 18) to generate adversarial images. The middle two columns change layers 7 to 12, and the last column uses the bottom 6 layers. We also show results on CelebA-HQ gender classification in the supplementary material. }
Figure \ref{fig:input-cond} illustrates adversarial examples conditioned on real input images using the procedure described in Section \ref{sec:input-cond}. Synthesized images resemble inputs with high fidelity, and set the initial values in our optimization process.  

\textcolor{black}{We also show results on semantic segmentation in Figure~\ref{fig:seg-results} in which we consider non-targeted attacks on DeepLab-v2 \cite{chen2017deeplab} with a generator trained on the COCO-stuff dataset \cite{caesar2018coco}. 
We iteratively modify modulation parameters at all layers, using a step size of $0.001$, to maximize the segmentation loss with respect to the given label map. 
 As we observe, subtle modifications to images lead to large drops in accuracy. } 
Object detection results and additional examples are provided in the supplementary material.  


\begin{table*}
\begin{center}
\begin{tabu}{ |[1.pt]c|[1.pt]c|c|[1.pt]c|c|[1.pt]c|c|[1.pt]c|c|[1.pt]c| }
\specialrule{.1em}{.05em}{.05em} 
 & \multicolumn{2}{c|[1.pt]}{Classification (LSUN)} & \multicolumn{2}{c|[1.pt]}{Classification (CelebA-HQ)} & \multicolumn{2}{c|[1.pt]}{Segmentation} & \multicolumn{2}{c|[1.pt]}{Detection} \\ 
 \specialrule{.1em}{.05em}{.05em} 
 & Clean & Adversarial & Clean & Adversarial & Clean & Adversarial & Clean & Adversarial \\ 
 \hline
 Adv. Trained & {\bf 89.5\%} & 78.4\% & {\bf 96.2\%} & 83.6\% & {\bf 69.1\%} & 60.2\% & {\bf 40.2\%} & 33.7\% \\
 \hline
 Original & 88.9\% & 0.0\% & 95.7\% & 0.0\% & 67.9\% & 2.7\% & 39.0\% & 2.0\% \\ 
\specialrule{.1em}{.05em}{.05em}
\end{tabu}
\end{center}
\vspace{-0.4cm}
\caption{Performance of adversarially trained and original models on clean and adversarial test images. Accuracy is shown for classification and segmentation, and Average Precision is shown for object detection.  
} \label{tab:adv-training} 
\vspace{-0.1cm}
\end{table*}

\begin{table*}
\begin{center}
\begin{tabu}{ |[1.pt]c|[1.pt]c|c|c|c|c|c|[1.pt]c|[1.pt]c| }
 \specialrule{.1em}{.05em}{.05em}
 Model & \multicolumn{6}{c|[1.pt]}{Attack} & Mean \\
 \specialrule{.1em}{.05em}{.05em} 
 & Clean & GAT & PGD & Spatial & Recolor & Perceptual & \\ 
 \specialrule{.1em}{.05em}{.05em}
 GAT (Ours) & {\bf 89.5\%} & 78.4\% & 39.4\% & 47.8\% & 52.3\% & 28.9\% & {\bf 42.1\%} \\
 \hline
 AT PGD \cite{madry2017towards} & 81.2\% & 6.3\% & 56.7\% & 5.1\% & 37.9\% & 2.8\% & 13.0\% \\
 \hline 
 AT AdvProp \cite{xie2020adversarial} & 89.4\% & 7.8\% & 57.6\% & 6.0\% & 38.5\% & 3.5\% & 22.7\% \\
 \hline
 AT Spatial \cite{xiao2018spatially} & 76.3\% & 5.4\% & 3.1\% & 66.0\% & 4.1\% & 2.2\% & 3.7\% \\
 \hline
 AT Recolor \cite{laidlaw2019functional} & 88.6\% & 4.7\% & 7.3\% & 0.4\% & 60.7\% & 1.7\% & 3.5\% \\
 \hline
 PAT \cite{laidlaw2020perceptual} & 72.4\% & 18.3\% & 40.1\% & 46.3\% & 42.5\% & 30.1\% & 36.5\% \\
 \specialrule{.1em}{.05em}{.05em}
\end{tabu}
\end{center}
\vspace{-0.4cm}
\caption{Accuracy of adversarially trained models against various attacks on the LSUN dataset. The mean accuracy of models on unseen attacks is shown in the last column.  
} \label{tab:unseen}
\vspace{-0.3cm}
\end{table*}

\subsection{Adversarial Training} 
Adversarial training increases robustness of models by injecting adversarial examples into training data. Adversarial training with norm-bounded examples degrades performance of the classifier on clean images as they have different underlying distributions.  
{We show that adversarial training with our unrestricted examples \textit{improves} the model's accuracy on clean test images as well as out-of-domain samples. 
To ensure that the model maximally benefits from these additional samples, we need to avoid unrealistic examples which do not resemble natural images. Therefore, we only include samples that can fool the model in less than a specific number of iterations. We use a threshold of $10$ as the maximum number of iterations, and demonstrate results on classification, semantic segmentation and object detection\footnote{In the supplementary material we show that limiting the number of iterations for norm-bounded perturbations is not effective for avoiding the accuracy drop on clean images.}. 
 We use the first $10$ 
 generated examples for each starting image in the segmentation and detection tasks.  
Table \ref{tab:adv-training} shows accuracy of the strengthened and original classifiers on clean and adversarial test images. 
For the segmentation and detection tasks we report the mean accuracy and average precision of adversarial images at iteration $10$. 
Similar to norm-constrained perturbations, adversarial training is an effective defense against our unrestricted attacks. 
Note that accuracy of the model on clean test images is improved after adversarial training. 
This is in contrast to training with norm-bounded adversarial inputs which hurts the classifier's performance on clean images, and  
it is due to the fact that unlike perturbation-based inputs, our generated images live on the manifold of realistic images as constrained by the generative model. 
}

We also evaluate the adversarially trained model against various unforeseen attacks to demonstrate generalizable robustness of the model. 
We consider several attacks including recoloring \cite{hosseini2018semantic,laidlaw2019functional}, spatial transformations \cite{xiao2018spatially}, perceptual \cite{laidlaw2020perceptual} and additive perturbations \cite{madry2017towards}. Results are shown in Table \ref{tab:unseen}, and are compared against other defense methods such as Adversarial Training with PGD (AT PGD) \cite{madry2017towards}, 
AT Spatial \cite{xiao2018spatially}, 
AT Recolor \cite{laidlaw2019functional}, PAT \cite{laidlaw2020perceptual} and 
AT AdvProp \cite{xie2020adversarial}. 
We observe that our adversarially trained model achieves superior robustness to these attacks. Unlike other methods which create a low-level change for each image, our approach generates a spectrum of low-level, mid-level and high-level changes around each sample, resulting in superior generalization to a variety of attacks without being trained against them. We also evaluate our attack against a certified defense in the supplementary material. 

Finally, we examine performance of the model on out-of-domain samples. We train a binary classifier on the two object classes of LSUN, i.e. cars and cats. We then evaluate the model on test samples from LSUN and ImageNet. 
Since several ImageNet classes represent cars and cats, we group all the relevant categories. Table \ref{tab:out-domain} demonstrates the results for regular and adversarially trained models. 
We observe that adversarial training with our examples improves generalization power of the model to LSUN test images as well as ImageNet out-of-domain samples.

\begin{table}
\begin{center}
\begin{tabu}{ |[1.pt]c|[1.pt]c|[1.pt]c|[1.pt] }
 \specialrule{.1em}{.05em}{.05em}
 & LSUN & ImageNet  \\
\specialrule{.1em}{.05em}{.05em}
Adv. Trained  & {\bf 94.7\%} 
& {\bf 92.0\%} 
\\
 \hline
Original & 94.2\% & 91.4\% \\
\specialrule{.1em}{.05em}{.05em}
\end{tabu}
\end{center}
\vspace{-0.5cm}
\caption{Generalization of the models to in-domain (LSUN) and out-of-domain (ImageNet) samples. 
} \label{tab:out-domain}
\vspace{-0.3cm}
\end{table}

\subsection{User Study} 
Norm-constrained attacks provide visual realism by $L_p$ proximity to a real input. 
To verify that our unrestricted adversarial examples are realistic and correctly classified by an oracle, we perform human evaluation using Amazon Mechanical Turk.   
In the first experiment, each adversarial image is assigned to three workers, and their majority vote is considered as the label. 
The user interface for each worker contains nine images,
and shows possible labels to choose from. 
We use $2400$ noise-based and $2400$ style-based adversarial images from the LSUN dataset, containing $200$ samples from each class ($10$ scene classes and $2$ object classes). The results indicate that $99.2\%$ of workers' majority votes match the ground-truth labels. This number is ${98.7\%}$ for style-based adversarial examples and ${99.7\%}$ for noise-based ones. As we observe in Figure \ref{fig:imgs}, noise-based examples do not deviate much from the original image, resulting in easier prediction by a human observer. On the other hand, style-based images show coarser changes, which in a few cases result in unrecognizable images or false predictions by the workers.      

We use a similar setup in the second experiment but for classifying real versus fake (generated). We also include $2400$ real images as well as $2400$ unperturbed images generated by Style-GAN. ${74.7}\%$ of unperturbed images are labeled by workers as real. This number is ${74.3}\%$ for noise-based adversarial examples and ${70.8\%}$ for style-based ones, 
indicating less than ${4\%}$ drop compared with unperturbed images generated by Style-GAN.  

\section{Conclusion and Future Work}

Existing works on adversarial defense assume a known threat model in advance. Therefore, adversaries can easily circumvent these defenses by using different types of attacks. This raises the need for defenses that are robust against unforeseen threat models. 
To this end, we incorporate diversity and realism in the examples used in adversarial training to bridge the distribution gap between real and adversarial examples. We leverage state-of-the-art generative models with disentangled representations which enable a range of low-level to high-level adversarial changes without leaving the manifold of natural images.
We demonstrate results on classification, segmentation and object detection tasks. We consider extending the model to other tasks and evaluating it against new threat models in the future.  



{\small
\bibliographystyle{ieee_fullname}
\bibliography{egbib}
}


\title{Supplementary Material: Robustness and Generalization via Generative Adversarial Training}

\date{}
\maketitle
\ificcvfinal\thispagestyle{empty}\fi



%



\section{Comparison with Song et al. \cite{song2018constructing}} 

We show that adversarial training with examples generated by [\textcolor{green}{33}] hurts the classifier's performance on clean images. Table \ref{tab:adv-song} demonstrates the results. We use the same classifier architectures as \cite{song2018constructing} and consider their basic attack.
We observe that the test accuracy on clean images drops by $1.3\%$, $1.4\%$ and $1.1\%$ on MNIST, SVHN and CelebA respectively. As we show in Table \textcolor{red}{1} of the main paper 
training with our examples improves the accuracy, demonstrating difference of our approach with that of \cite{song2018constructing}.   

\begin{table*}[h]
\begin{center}
\begin{tabu}{ |[1.pt]c|[1.pt]c|c|[1.pt]c|c|[1.pt]c|c|[1.pt]c| }
\specialrule{.1em}{.05em}{.05em} 
 & \multicolumn{2}{c|[1.pt]}{MNIST} & \multicolumn{2}{c|[1.pt]}{SVHN} & \multicolumn{2}{c|[1.pt]}{CelebA} \\
 \specialrule{.1em}{.05em}{.05em} 
 & Clean & Adversarial & Clean & Adversarial & Clean & Adversarial \\ 
 \hline

 Adv. Trained & 98.2\% & 84.5\% & 96.4\% & 86.4\% & 96.9\% & 85.9\% \\
\hline
Original & 99.5\% & 12.8\% & 97.8\% & 14.9\% & 98.0\% & 16.2\% \\ 
\specialrule{.1em}{.05em}{.05em}
\end{tabu}
\end{center}
\caption{Accuracy of adversarially trained and original models on clean and adversarial test images from \cite{song2018constructing}. 
} \label{tab:adv-song} 
\end{table*}

\begin{figure*}
\centering
\small
\setlength{\tabcolsep}{1pt}
\begin{tabular}{cc}
  \includegraphics[width=.48\textwidth]{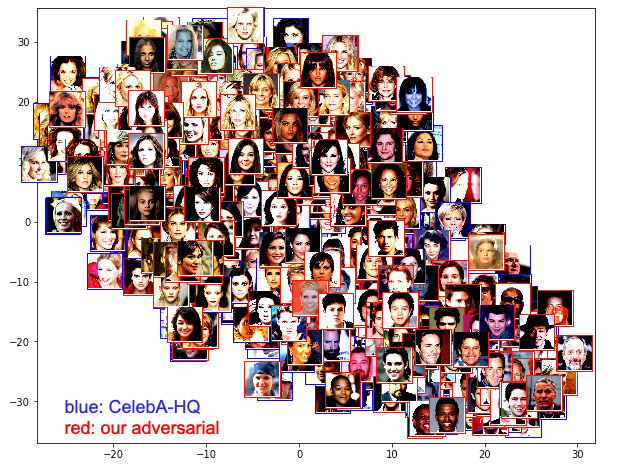} &
  \includegraphics[width=.48\textwidth]{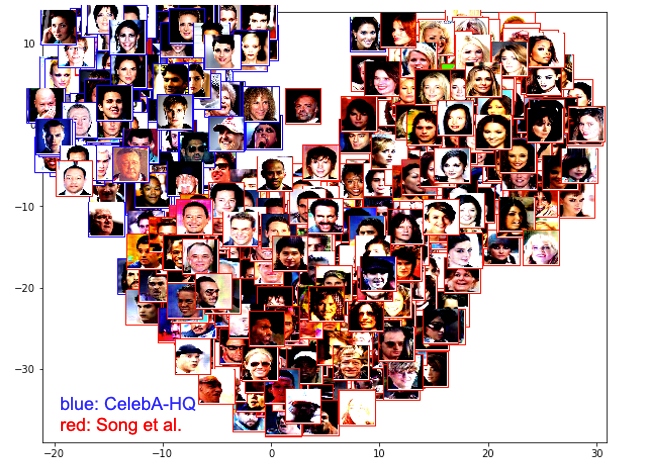} \\
\end{tabular}
\caption{t-SNE plot comparing distributions of real images with adversarial examples from our approach and Song et al.}
\label{fig:t_sne}
\end{figure*}

To further illustrate and compare distributions of real and adversarial images, we use a pre-trained VGG network to extract features of each image from CelebA-HQ, our adversarial examples, and those of \cite{song2018constructing}, and then plot them with t-SNE embeddings as shown in Figure \ref{fig:t_sne}. We can see that the embeddings of CelebA-HQ real and our adversarial images are blended while those of CelebA-HQ and Song et al.'s adversarial examples are more segregated. 
This again provides evidence that our adversarial images stay closer to the original manifold and hence could be more useful as adversarial training data. 

\section{Number of Iterations}
 To make sure the iterative process always converges in a reasonable number of steps, we measure the number of updates required to fool the classifier on $1000$ randomly-selected images. Results are shown in Table~\ref{tab:iterations}. 
 Note that for targeted attacks we first randomly sample a target class different from the ground-truth label for each image. 

\begin{table}[h]
\begin{center}
\begin{tabu}{ |[1.pt]c|[1.pt]c|c|[1.pt]c|[1.pt] }
 \specialrule{.1em}{.05em}{.05em}
 &  \multicolumn{2}{c|[1.pt]}{LSUN} & {CelebA-HQ} \\ 
 \specialrule{.1em}{.05em}{.05em}
 & Targeted & Non-targeted & \\
 \hline
 Style-based & ${9.1 \pm 4.2}$ & ${6.8 \pm 3.6}$ & $7.3 \pm 3.0$ \\
 \hline
 Noise-based & $4.5 \pm 1.7$ & ${3.7 \pm 1.8}$ & $6.2 \pm 4.1$ \\
\specialrule{.1em}{.05em}{.05em}
\end{tabu}
\end{center}
\caption{Average number of iterations (mean $\pm$ std) required to fool the classifier. 
} \label{tab:iterations}
\end{table}


\section{Evaluation on Certified Defenses} 
Adversaries can circumvent defenses tailored for a specific type of attack by using new threat models. To demonstrate this, we evaluate our attack on a certified defense against norm-bounded perturbations. Cohen et al. \cite{cohen2019certified} propose a certified defense using randomized smoothing with Gaussian noise, which guarantees a certain top-1 accuracy for perturbations with $L_2$ norm less than a specific threshold. 
We use ${400}$ noise-based and ${400}$ style-based adversarial images from the object categories of LSUN. 
Our adversarial examples are evaluated against a randomized smoothing classifier based on ResNet-50 using Gaussian noise with standard deviation of $0.5$. 
Table \ref{tab:certified} shows accuracy of the model on clean and adversarial images. As we observe, the accuracy drops on adversarial inputs, and the certified defense is not  effective against our attack. Note that we stop updating adversarial images as soon as the model is fooled. If we keep updating for more iterations afterwards, we can achieve even stronger attacks.   

\vspace{-0.1cm}
\begin{table}[h]
\begin{center}
\begin{tabular}{ |c|c|c| }
\hline
 & Accuracy  \\ 
 \hline
  Clean &  63.1\% \\
 \hline
 Adversarial (style) &  21.7\%  \\
 \hline
 Adversarial (noise) & 37.8\%  \\
 \hline
\end{tabular}
\end{center}
\vspace{-0.3cm}
\caption{Accuracy of a certified classifier equipped with randomized smoothing on our adversarial images. 
} \label{tab:certified}
\vspace{-0.3cm}
\end{table}

\section{Object Detection Results}
\begin{figure*}
\begin{center}
\includegraphics[width=\textwidth]{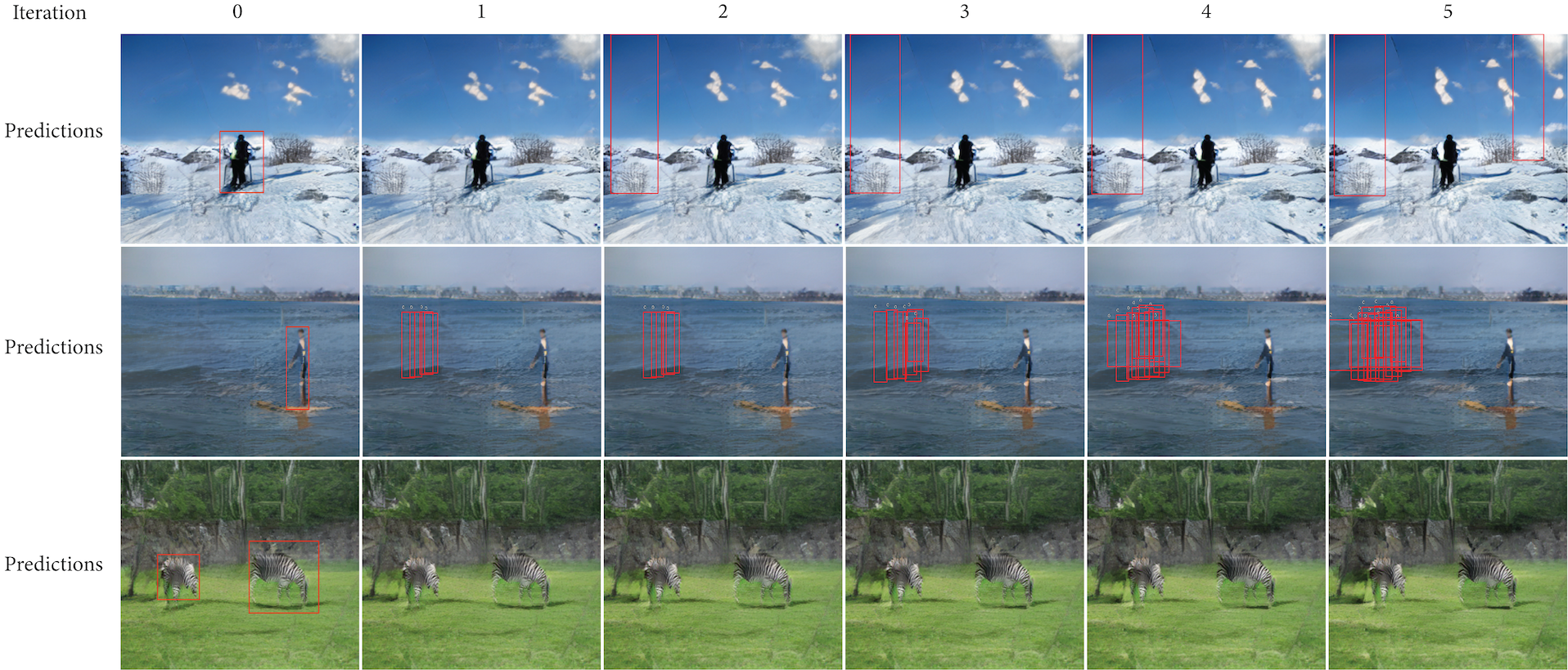}
\end{center}
  \caption{Unrestricted adversarial examples for object detection. Generated images and their corresponding predictions are shown for different number of iterations. 
  }
\label{fig:adv-obj}
\end{figure*}

Figure \ref{fig:adv-obj} illustrates results on the object detection task using the RetinaNet target model [\textcolor{green}{28}]. We observe that small changes in the images lead to incorrect bounding boxes and predictions by the model.  


\section{Impact of $\gamma$ and $\beta$ on Semantic Segmentation}

\begin{figure*}
\begin{center}
\includegraphics[width=1.0\linewidth]{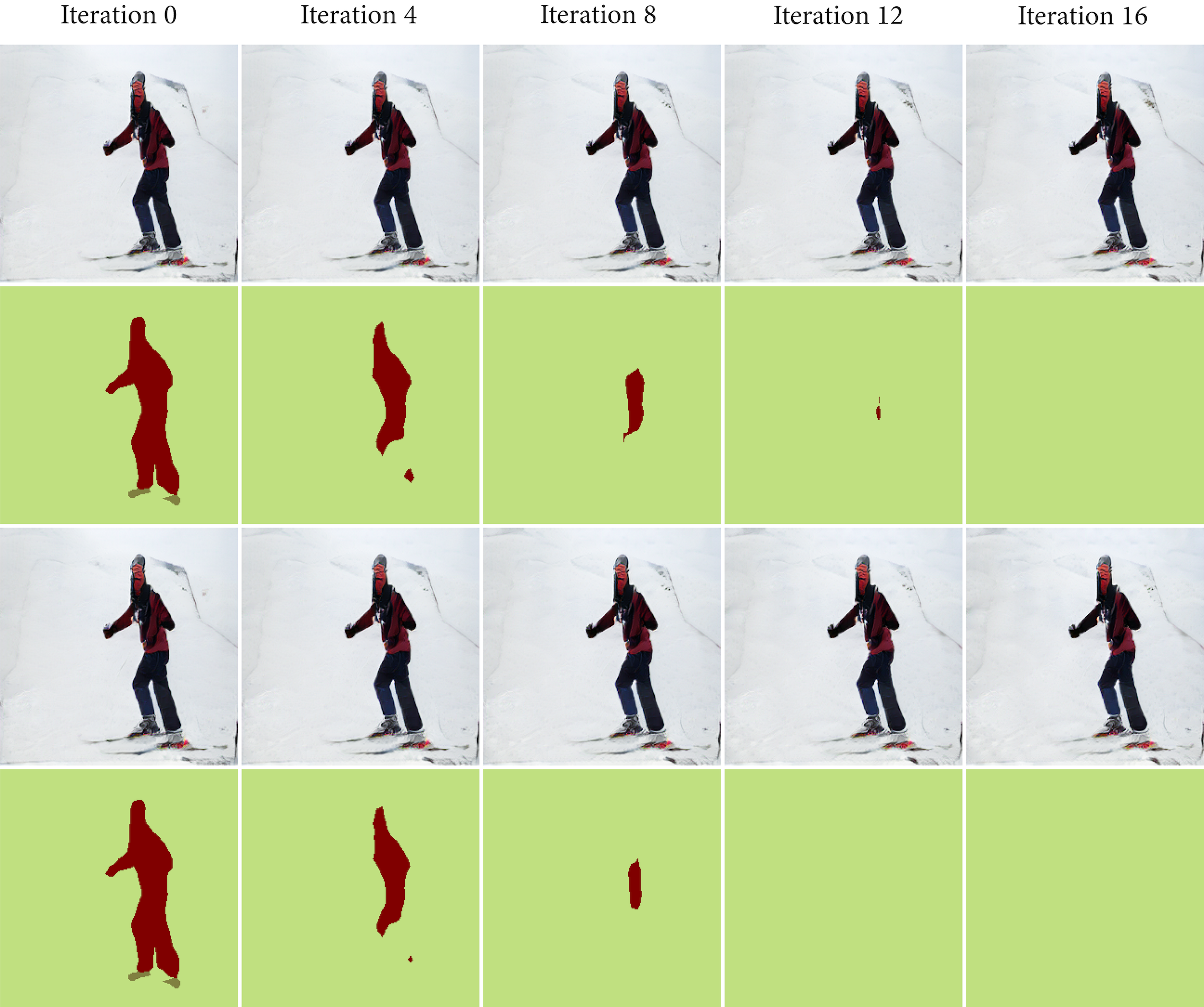}
\end{center} 
  \caption{Impact of separately modifying $\gamma$ and $\beta$ parameters on segmentation results. Modified images at different iterations and corresponding predictions are shown. 
  In the first two rows only the $\gamma$ values are changed and in the last two rows only the $\beta$ values are modified.  
  }
\label{fig:gamma-beta}
\end{figure*}

In segmentation results shown in Figure \ref{fig:adv-seg} we simultaneously modify both $\gamma$ and $\beta$ parameters of the SPADE module. We can also consider the impact of modifying each parameter separately. Figure \ref{fig:gamma-beta} illustrates the results. As we observe, changing $\gamma$ and $\beta$ modifies fine details of the images which are barely perceptible yet they lead to large changes in predictions of the segmentation model.

\section{Experiments on other datasets}

While we show results on the categories that StyleGAN is originally trained on, our approach and StyleGAN can also be trained on other datasets such as CIFAR and ImageNet\footnote{Implementations of StyleGAN2 and StyleGAN2-ADA on CIFAR and ImageNet are provided in https://github.com/justinpinkney/awesome-pretrained-stylegan2 and https://github.com/NVlabs/stylegan2-ada}.  
We provide results on the CIFAR-10 dataset in the following using the Wide ResNet classifier:

\begin{center}
\begin{tabular}{|c|c|c|} 
\hline
  & Clean & Adversarial  \\ 
\hline
 Adv. Trained & 96.5\% & 76.4\% \\ 
\hline
 Original & 96.1\% & 0.0\% \\ 
\hline
\end{tabular}
\vspace{+0.2cm}
\end{center}

\section{Adversarial Changes to Single Images}
\begin{figure*}
\begin{center}
\includegraphics[width=0.95\textwidth]{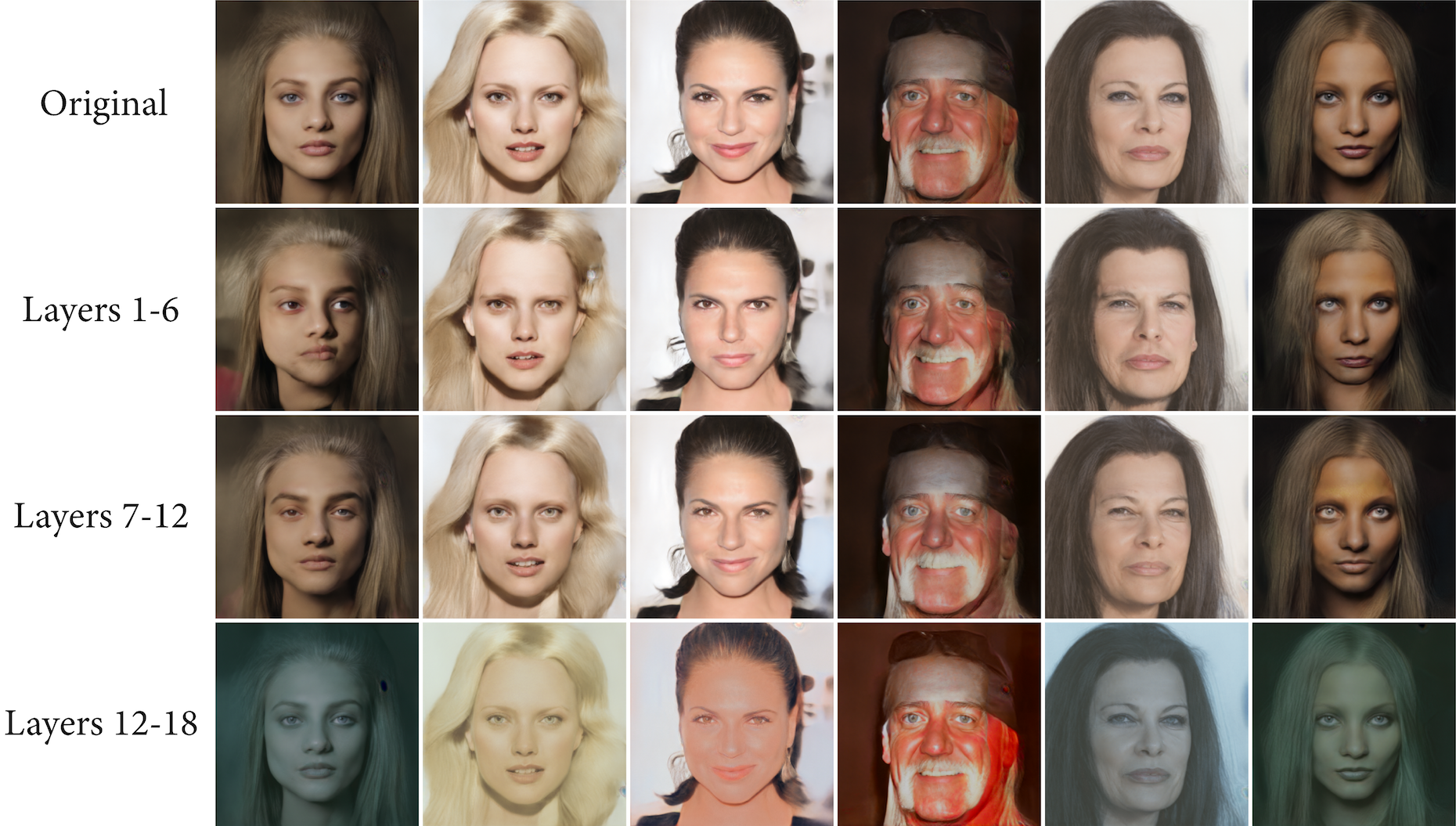}
\end{center} 
  \caption{Impact of manipulating different layers of the network on generated adversarial images. 
  }
\label{fig:single}
\end{figure*}

{Figure \ref{fig:single} illustrates how images vary as we manipulate specific layers of the network. We observe that each set of layers creates different adversarial changes. For instance, layers 12 to 18 mainly change low-level color details. }

\section{Adversarial Training with Norm-bounded Perturbations}

We consider adversarial training with norm-bounded perturbations and limit the number of iterations to make the setup comparable with our unrestricted adversarial training. 
Specifically, we use Iterative-FGSM with $\epsilon=4$ and a bounded number of steps. Results are shown in Table \ref{tab:adv-norm}. Note that accuracy of the models drop on clean images although we use a weak attack. This is in contrast to training with our unrestricted adversarial examples that improves the accuracy.

\begin{table}[h]
\begin{center}
\begin{tabu}{ |[1.pt]c|[1.pt]c|c|c|[1.pt]c|}
\specialrule{.1em}{.05em}{.05em}
 & & \multicolumn{1}{c|}{IFGSM-2} & \multicolumn{1}{c|[1.pt]}{IFGSM-5} \\
\specialrule{.1em}{.05em}{.05em}
 & Original & Adv. Trained & Adv. Trained \\
 \hline
 LSUN & 88.9\% & 88.4\% & 87.8\% \\
\hline
 CelebA-HQ & 95.7\% & 95.1\% & 94.6\% \\ 
\specialrule{.1em}{.05em}{.05em}
\end{tabu}
\end{center}
\caption{Adversarial Training with norm-bounded perturbations. Iterative-FGSM ($\epsilon=4$) with a maximum of 2 and 5 iterations is considered, and accuracy of the adversarially trained and original models on clean test images are shown. } 
\label{tab:adv-norm} 
\end{table}


\section{Additional Examples}

\begin{figure*}
\begin{center}
\includegraphics[width=\textwidth]{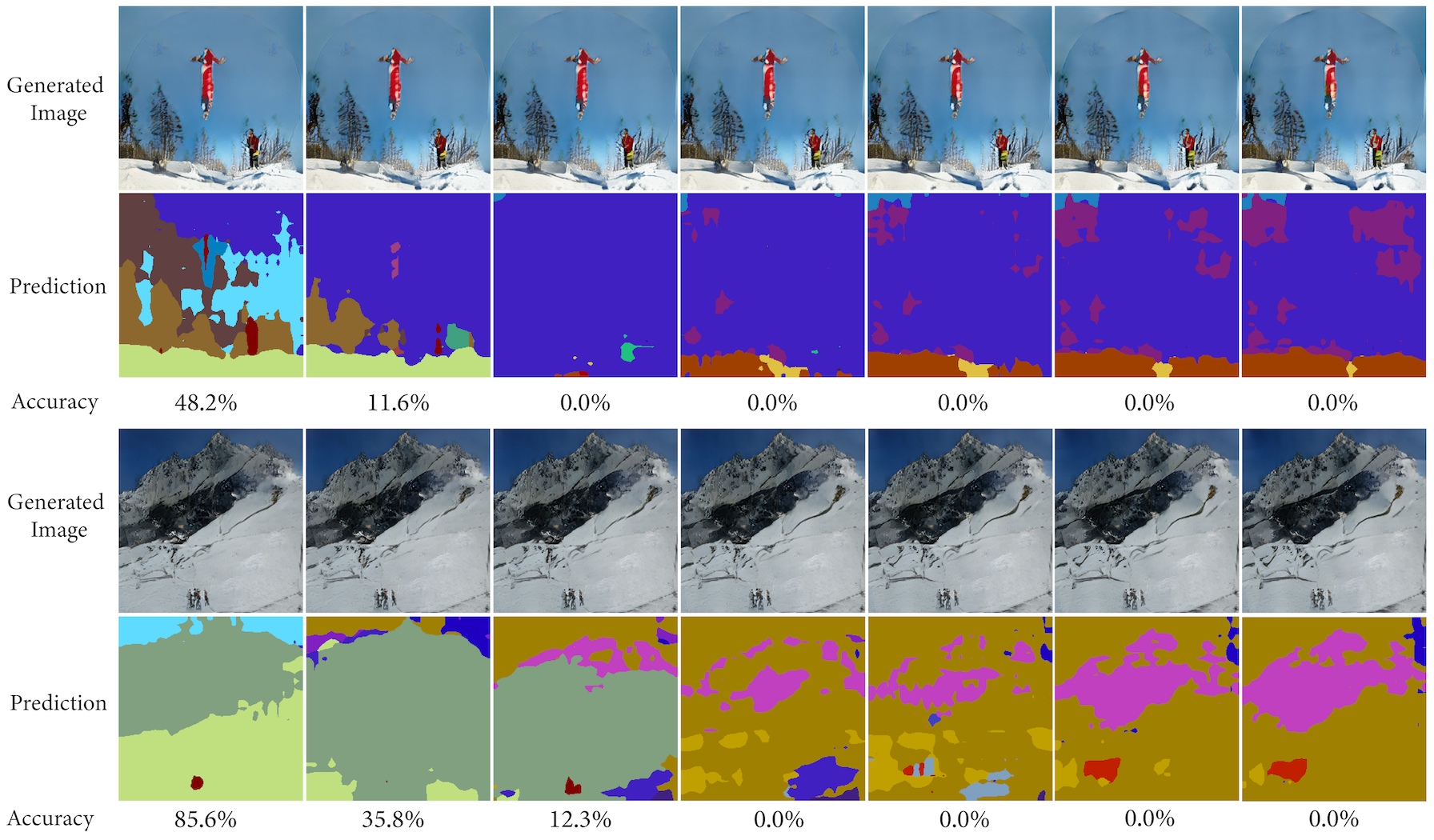}
\end{center}
  \caption{Unrestricted adversarial examples for semantic segmentation. Generated images, corresponding predictions and their accuracy (ratio of correctly predicted pixels) are shown for different number of iterations. 
  }
\label{fig:adv-seg}
\end{figure*}

\begin{figure*}
\begin{center}
\includegraphics[width=\textwidth]{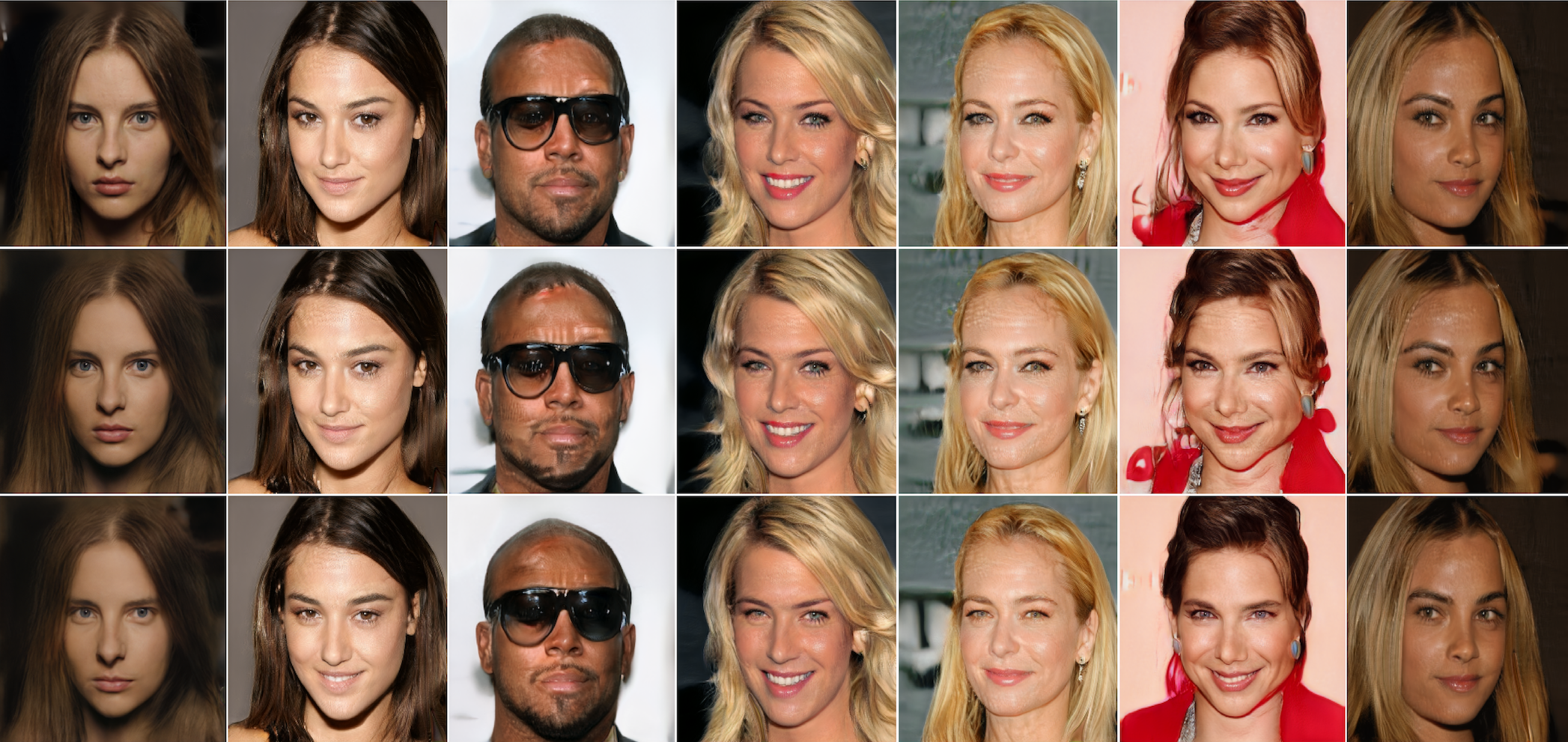}
\end{center}
  \caption{Unrestricted adversarial examples on CelebA-HQ gender classification. From top to bottom: Original, noise-based and style-based adversarial images. Males are classified as females and vice versa.
  }
\label{fig:celeba-add}
\end{figure*}

\begin{figure*}
\begin{center}
\begin{subfigure}{1.0\textwidth}
\includegraphics[width=\textwidth]{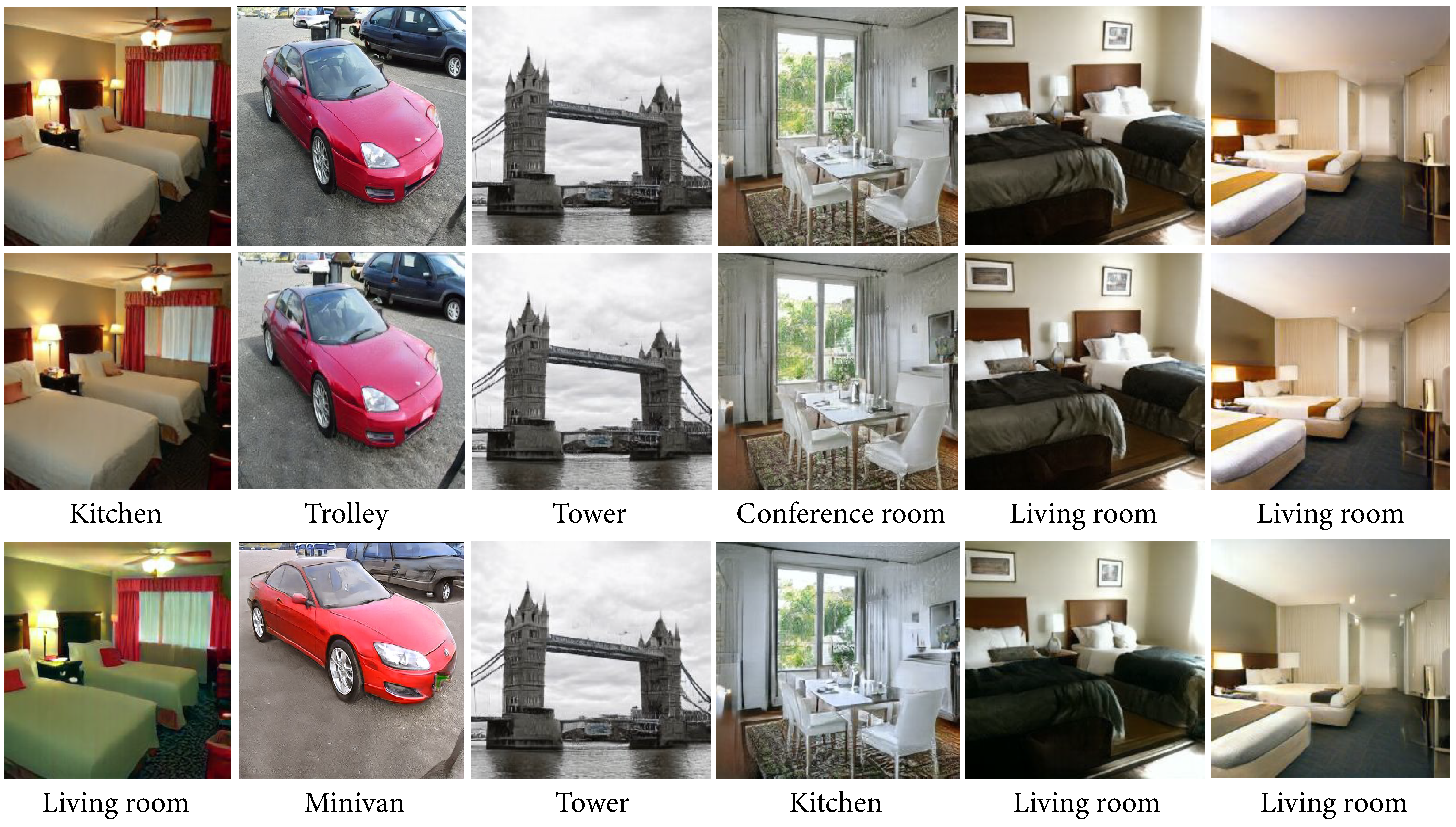}
\caption{Non-targeted}
\end{subfigure}
\begin{subfigure}{1.0\textwidth}
\includegraphics[width=\textwidth]{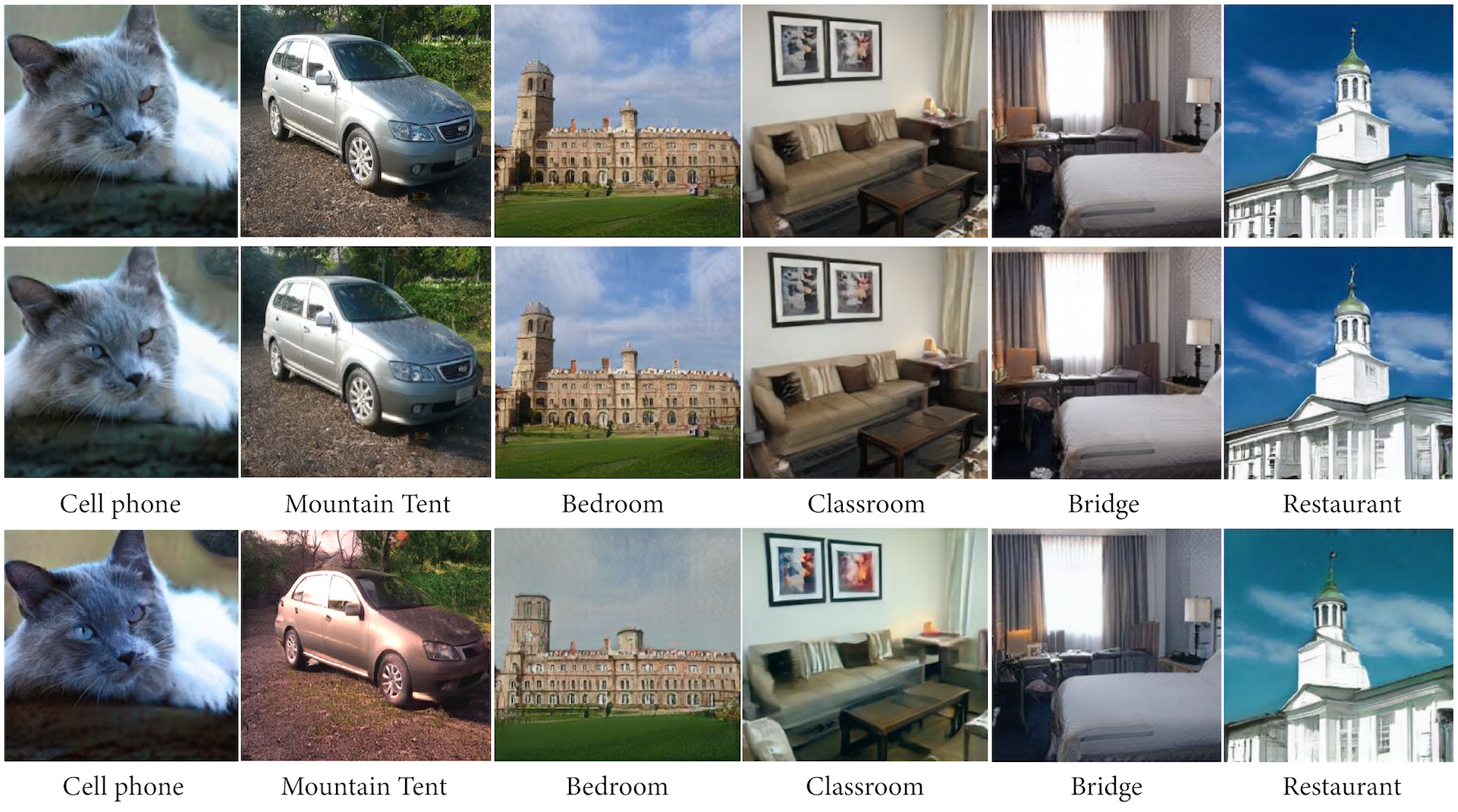}
\caption{Targeted}
\end{subfigure}
\end{center}
\vspace{-0.3cm}
  \caption{Unrestricted adversarial examples on LSUN for a) non-targeted and b) targeted attacks. From top to bottom: original, noise-based and style-based images. 
  }
\label{fig:lsun-add}
\end{figure*}

We also provide additional examples and higher-resolution images in the following. Figure \ref{fig:adv-seg} depicts additional examples on the segmentation task. Figure \ref{fig:celeba-add} illustrates adversarial examples on CelebA-HQ gender classification, and Figure \ref{fig:lsun-add} shows additional examples on the LSUN dataset. 
Higher-resolution versions for some of the adversarial images are shown in Figure \ref{fig:high-res}, which particularly helps to distinguish subtle differences between original and noise-based images.

\begin{figure*}[t]
\begin{center}
\includegraphics[width=0.93\textwidth]{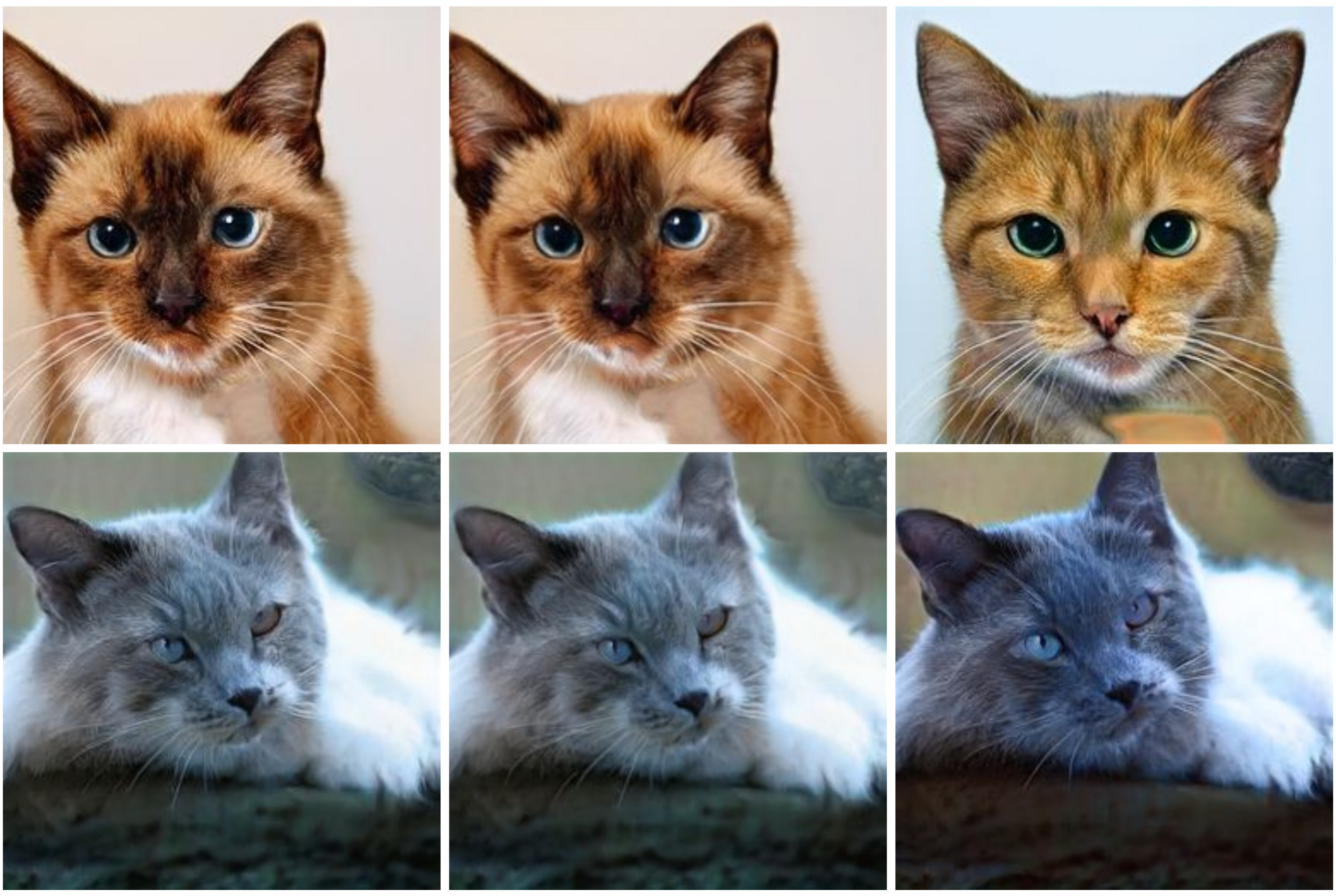}
\includegraphics[width=0.93\textwidth]{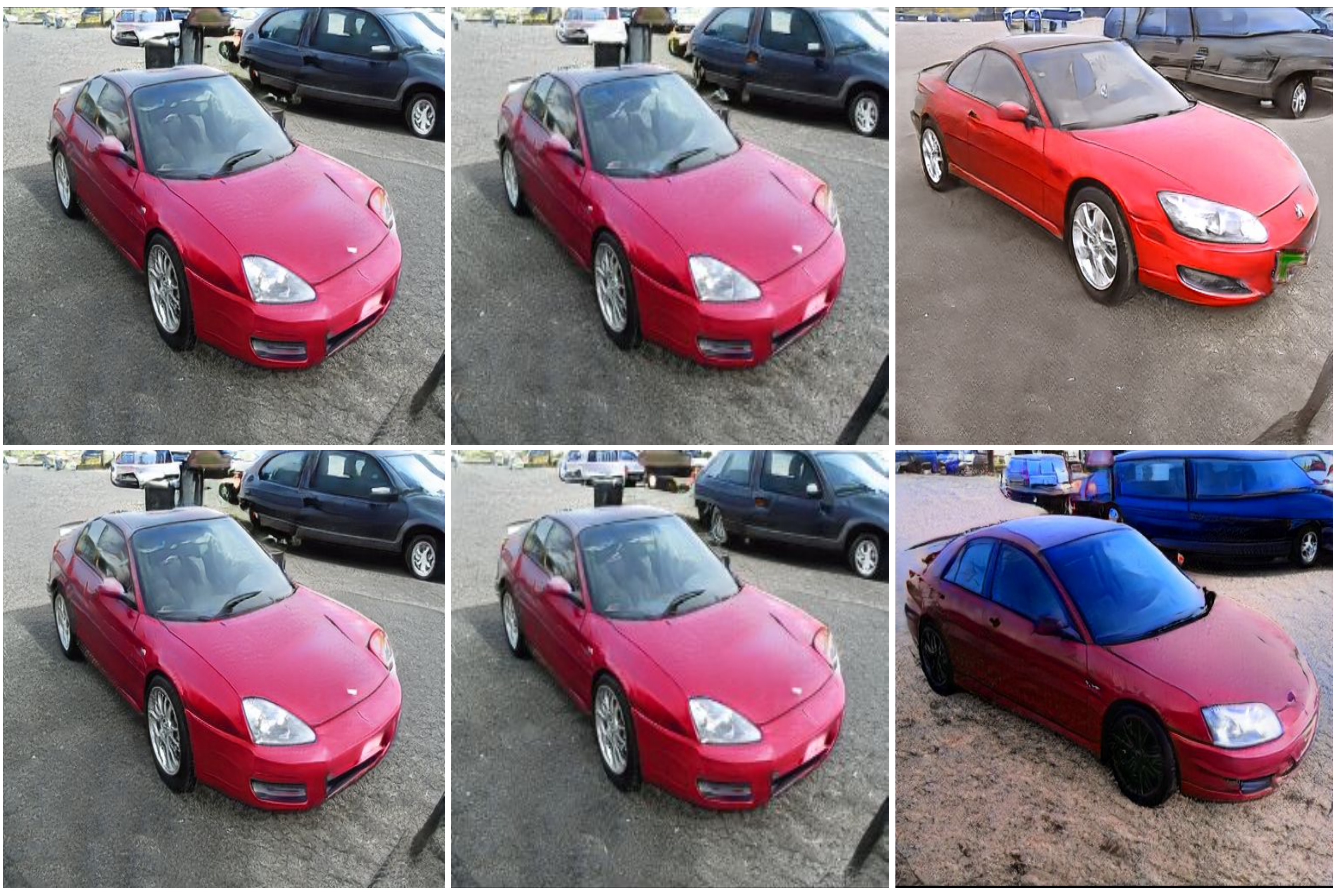}
\end{center}
\vspace{-0.3cm}
  \caption{High resolution versions of adversarial images. From left to right: original, noise-based and style-based images.
  }
\label{fig:high-res}
\end{figure*}

\begin{figure*}\ContinuedFloat
\begin{center}
\includegraphics[width=0.93\textwidth]{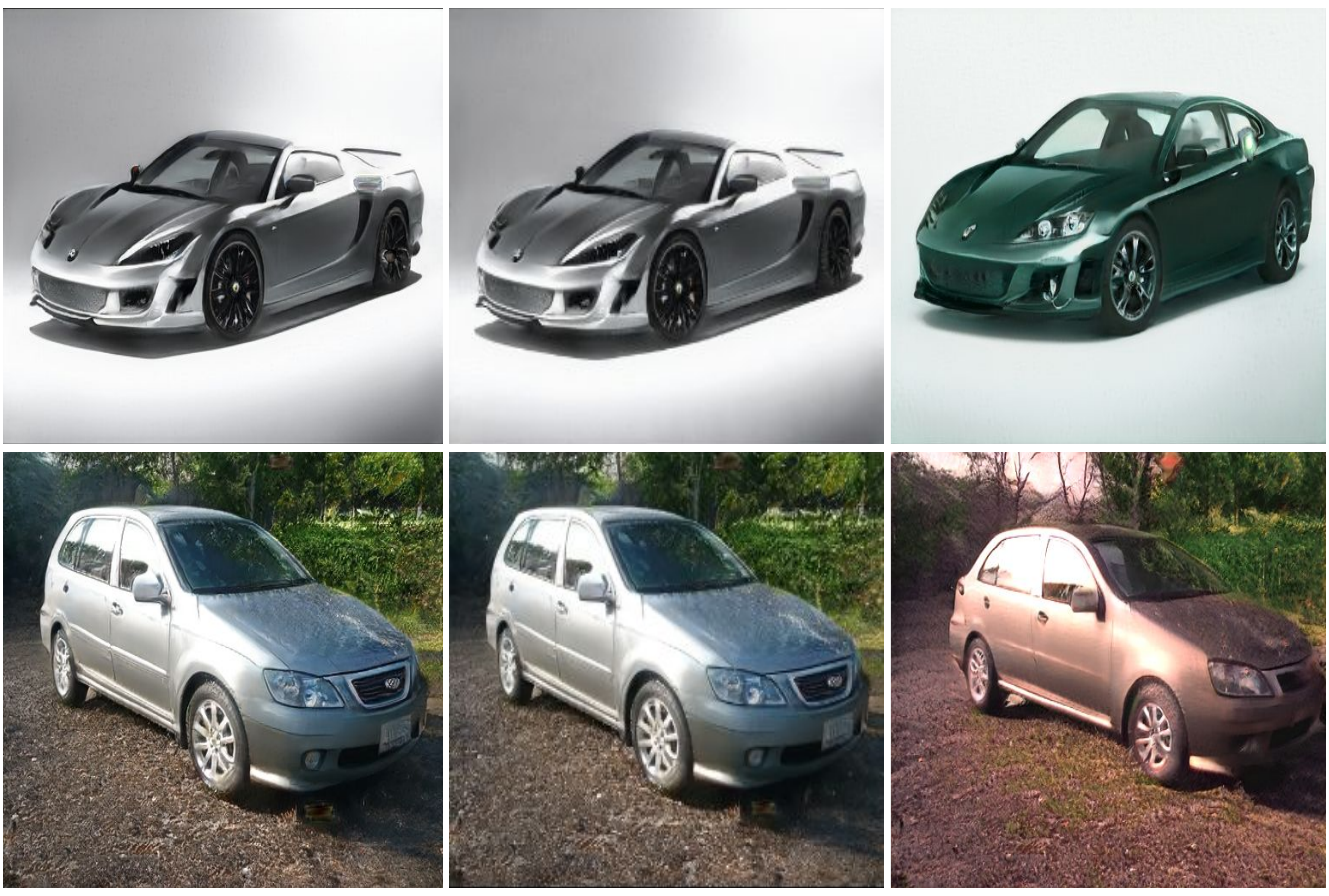}
\includegraphics[width=0.93\textwidth]{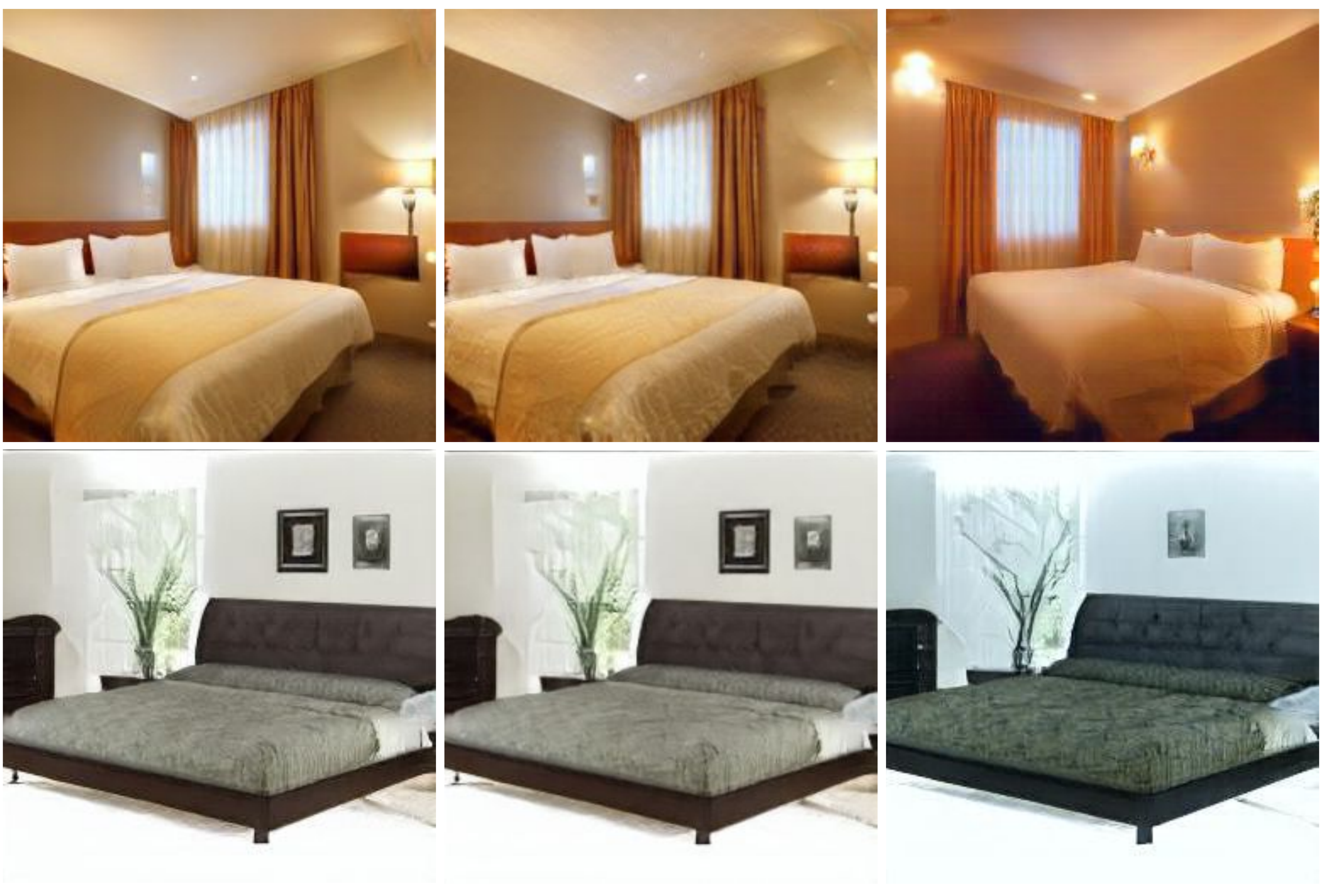}
\end{center}
\vspace{-0.4cm}
\caption{(cont.) High resolution versions of adversarial examples. From left to right: original, noise-based and style-based images.}
\label{fig:high-res}
\end{figure*}

\end{document}